\documentclass{article}

\newcommand{\BH}[1]{\textcolor[rgb]{0,0,1}{\textbf{#1}}}
\newcommand{\RH}[1]{\textcolor[rgb]{1,0,0}{\textbf{#1}}}

\usepackage[square,numbers]{natbib}
\usepackage[main, final]{neurips_2026}
\usepackage{graphicx}
\usepackage{refcount}
\usepackage{color}
\usepackage{makecell}
\usepackage{multirow}
\usepackage{amsmath}
\usepackage[utf8]{inputenc} 
\usepackage[T1]{fontenc}    
\usepackage{hyperref}       
\usepackage{url}            
\usepackage{booktabs}       
\usepackage{amsfonts}       
\usepackage{nicefrac}       
\usepackage{microtype}      
\usepackage{xcolor}         

\title{V2VFusion: Text-Controlled Video-to-Video Diffusion for Degradation-Aware Video Fusion}

\author{
  Jiajun Chen$^1$,\,
  Han Xu$^{1}$\thanks{Co-corresponding authors.} ,\,
  Yunfei Hang$^1$,\,
  Zhuoya Zhang$^1$,\,
  Guangcan Liu$^1$,\,
  Jiayi Ma$^{2}$\footnotemark[1]\\[2pt]
  $^1$ School of Automation, Southeast University, Nanjing, China\\
  $^2$ School of Robotics, Wuhan University, Wuhan, China\\[2pt]
  \texttt{\{jiajunchen,xu\_han,huang\_yunfei,zhuoyazhang,guangcanliu\}@seu.edu.cn}\\
  \texttt{jyma2010@gmail.com}
}

\begin{document}

\maketitle

\begin{abstract}
Video fusion integrates complementary information in multiple source video streams into a single sequence for complete scene perception. Existing methods perform frame-wise fusion and lack unified spatio-temporal modeling across sequences, damaging temporal coherence across sequences. When dealing with degradations, they also ignore temporal dynamics and overload a single network with diverse degradations, leading to limited and imbalanced degradation suppression. To address these issues, we propose V2VFusion, the first unified video-to-video fusion framework that directly generates temporally coherent, high-quality fused videos from degraded multi-source inputs. By reformulating video fusion as a video-to-video conditional generation problem in latent diffusion space, our approach holistically models spatio-temporal dependencies across entire sequences and intrinsically ensures sequence-level consistency. On this basis, a text-guided degradation-conditioned ControlNet bridges high-level textual degradation descriptions with low-level visual priors, enabling interpretable and controllable guidance for structure-preserving fusion beyond purely data-driven conditioning. To handle heterogeneous and compound degradations, a state-modulated degradation-aware hierarchical mixture-of-experts module decouples expert selection into diffusion-state-aware filtering and degradation-aware routing. It prevents unreliable routing and fosters expert specialization to mitigate representation conflicts among diverse degradations. Experiments on infrared-visible, multi-exposure, and multi-focus video fusion tasks and complex degradations validate that V2VFusion surpasses existing methods in fusion quality and temporal stability. The code is available at \href{https://github.com/ChenDoubleJ/V2VFusion}{\texttt{https://github.com/ChenDoubleJ/V2VFusion}}.
\end{abstract}

\section{Introduction}
Real-world visual perception is inherently acquired in continuous video streams rather than isolated images \cite{ding2026cmvf}. Due to sensing limitations, video captured by a single modality/setting often suffers from incomplete scene perception. Thus, video fusion aims to integrate complementary information from multiple video sequences while maintaining temporal consistency, enabling robust and informative visual understanding. It plays a critical role in numerous downstream applications, \textit{e.g.}, unmanned aerial surveillance \cite{hu2022disguised}, autonomous driving \cite{muller2009cognitively, wen2024panacea}, intelligent monitoring systems \cite{liu2013intelligent, liu2022target}.

\begin{figure}[!t]
    \centering
    \includegraphics[width=1.0\linewidth]{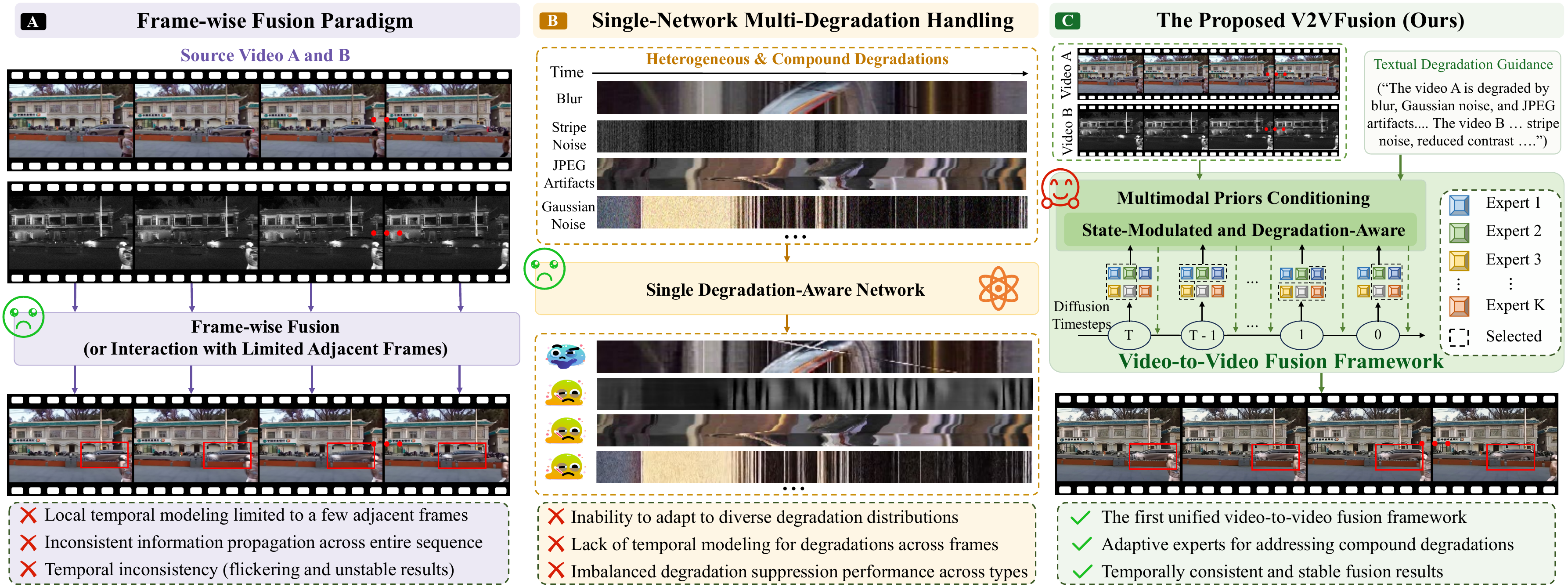}
    \vspace{-0.10in}
    \caption{Existing video fusion paradigm. The fusion result in A is generated by VideoFusion \cite{tang2025videofusion}.}
    \label{intro}
\end{figure}

Existing fusion research has largely been confined to static image fusion. It naturally lays the tractable and foundational groundwork for handling video streams. Image fusion achieves significant progress with methods based on
auto-encoder~\cite{li2018densefuse, li2020nestfuse}, CNN~\cite{xu2020u2fusion, xu2020fusiondn, cao2023multi}, Transformer~\cite{ma2022swinfusion, rao2023tgfuse,xu2026iterrf}, GAN~\cite{ma2019fusiongan, ma2020ddcgan}, and diffusion models~\cite{yue2023dif, yi2024diff, li2025towards, xu2026diff}. While these approaches demonstrate impressive performance in single-frame image fusion, they lack explicit temporal modeling. When directly extended to video fusion, they often fail to maintain temporal continuity, leading to flickering artifacts and motion inconsistencies in fused videos, as shown in Fig.~\ref{intro}.

Recently, several video fusion methods have been proposed~\cite{zhao2025unified,gong2025temcoco,tang2025videofusion}. 
They follow a video-to-frame-to-video paradigm: source videos are first decomposed into short frame sequences. Rather than directly generating a temporally coherent fused video as a unified output, this paradigm generates temporally independent fused frames, which are reassembled into a fused video. Thus, temporal dependencies are only implicitly propagated through neighboring frames, resulting in short-range interactions and inconsistent information propagation across sequence. As a result, temporal inconsistencies accumulate over time, still producing flickering artifacts in fused videos. 
Although temporal consistency or optical flow-based constraints are introduced, they remain external corrections rather than a fundamental solution to the underlying frame-wise modeling paradigm. Thus, it remains challenging and pressing to design a unified video-level spatial–temporal modeling rather than frame-wise processing to produce temporally coherent and sequence-stable fusion results.

Moreover, due to imperfect imaging conditions or transmission processes, real-world video sequences are often affected by various degradations, such as noise, blur, and JPEG compression artifacts. 
Unlike static images, video degradations exhibit complex temporal behaviors. Some degradations persist across consecutive frames, while others fluctuate independently across frames, making degradation removal in video fusion more challenging. (1) While some degradation-aware image fusion methods~\cite{yi2024text, zhang2024text, xu2025deno, xu2025urfusion, cao2025mmaif} can deal with single-frame image degradation, they operate on individual frames independently. They fail to exploit the structural consistency across consecutive frames and ignore the temporal degradation evolution, leading to inconsistent suppression and temporal artifacts when directly applied to video with compound degradations. (2) While some data-driven video fusion methods~\cite{gong2025temcoco,tang2025videofusion} can handle degraded inputs, their single network is unable to cope well in complex scenarios with compound degradations. Different degradations are heterogeneous in nature and affect contents differently. A single network cannot learn disentangled degradation-specific representations and effectively model heterogeneous degradations, making it struggle with complex degradation mixtures. Thus, designing a framework that models temporal structures while adaptively handling heterogeneous and compound degradations remains challenging.

To address these challenges, we propose V2VFusion, a novel text-controlled video-to-video diffusion framework for high-quality video fusion. Fundamentally departing from conventional frame-wise pipelines, we for the first time reformulate video fusion as a direct video-to-video conditional generation problem, enabling unified spatio-temporal modeling over entire video sequences and producing temporally coherent fused videos.
Within this framework, we introduce a text-guided degradation-conditioned ControlNet that unifies multi-modal priors (i.e., degraded video and textual degradation description), bridging low-level visual cues with high-level language understanding to adaptively regulate restoration and fusion. This provides a flexible and interpretable control mechanism beyond purely data-driven conditioning.
To further handle heterogeneous and compound degradations, we propose a state-modulated degradation-aware hierarchical mixture-of-experts module. Unlike conventional MoE that routes experts solely based on input features, our design decomposes expert selection into a hierarchical process conditioned on both the diffusion state and the degradation type. It avoids unreliable expert activation in denoising stages and encourages experts to specialize in distinct degradation patterns, mitigating representation conflicts under degradation mixtures. Extensive experiments on multiple video fusion tasks under diverse degradations demonstrate that V2VFusion consistently outperforms existing methods. Our main contributions are summarized as follows:

\textbullet \hspace{0.03in} We propose the first video-to-video fusion framework that directly generates fused videos rather than relying on frame-wise fusion paradigms. By reformulating video fusion as a conditional video-to-video generation problem in the latent diffusion space, it holistically models spatio‑temporal dependencies across entire sequences, intrinsically ensuring temporally coherent fusion results.

\textbullet \hspace{0.03in} We develop a text‑guided degradation‑conditioned ControlNet that bridges high‑level semantic descriptions with low‑level visual priors within the latent diffusion process. By unifying these multi-modal priors, it enhances degradation understanding and enables structure‑preserving generation through interpretable and controllable guidance, moving beyond purely data‑driven fusion paradigms.


\textbullet \hspace{0.03in} We propose a state-modulated degradation-aware hierarchical mixture-of-experts module to handle heterogeneous and compound degradations in video fusion. Instead of the conventional MoE that uniformly routes experts solely based on input features, it decouples expert selection into diffusion‑state‑aware filtering and degradation‑aware routing. This hierarchical design prevents unreliable routing across different denoising stages and fosters expert specialization, thereby mitigating representation conflicts among heterogeneous degradations.

\textbullet \hspace{0.03in} We train a unified fusion model for multiple video fusion tasks, including infrared–visible, multi-exposure, and multi-focus fusion. Extensive experiments show its consistent performance across these tasks and robust fusion quality under diverse real-world degradations such as blur, Gaussian noise, stripe noise, and JPEG compression artifacts.

\section{Related Work}
\textbf{Image Fusion}. 
Early deep learning approaches typically employ auto-encoder~\cite{li2018densefuse, li2020nestfuse}. 
Subsequent studies have explored CNN~\cite{xu2020u2fusion, xu2020fusiondn, cao2023multi, chen2026breaking} and Transformer~\cite{ma2022swinfusion, rao2023tgfuse} architectures, learning fusion strategies in a fully data-driven manner. Furthermore, by incorporating adversarial learning, generative adversarial networks (GANs) have also been introduced to image fusion~\cite{ma2019fusiongan, ma2020ddcgan} while they still suffer from unstable training and potential mode collapse. Thus, diffusion-based models \cite{yue2023dif, yi2024diff, li2025towards} have emerged as a promising alternative, demonstrating strong generative capability and more stable training behavior. 
However, these methods assume ideal input quality and overlook real-world degradations. While degradation-aware methods~\cite{tang2024drmf, zhang2024text, xu2025deno, xu2025urfusion, cao2025mmaif} address this, they are still
image-level and ignore temporal video characteristics.
\newline
\textbf{Video Fusion}. Compared with image fusion, video fusion requires modeling temporal consistency across frames. Recent studies aim to address this challenge by introducing temporal alignment or optical flow estimation to exploit cross-frame information. For example, TemCoCo~\cite{gong2025temcoco} enhances temporal consistency in video fusion by introducing visual–semantic collaboration and temporal cooperative learning. VideoFusion~\cite{tang2025videofusion} builds a spatio-temporal collaborative network to jointly exploit cross-modal complementarity and temporal dependencies. UniVF~\cite{zhao2025unified} proposes a unified multi-frame learning framework that aggregates neighboring frames for general video fusion. However, they focus on cross-modal information aggregation and typically assume relatively high-quality input videos. When source videos suffer from complex degradations, their performance often degrades.

\section{Proposed Method}

\subsection{Overall Architecture}
\begin{figure*}[!t]
    \centering    \includegraphics[width=1.0\linewidth]{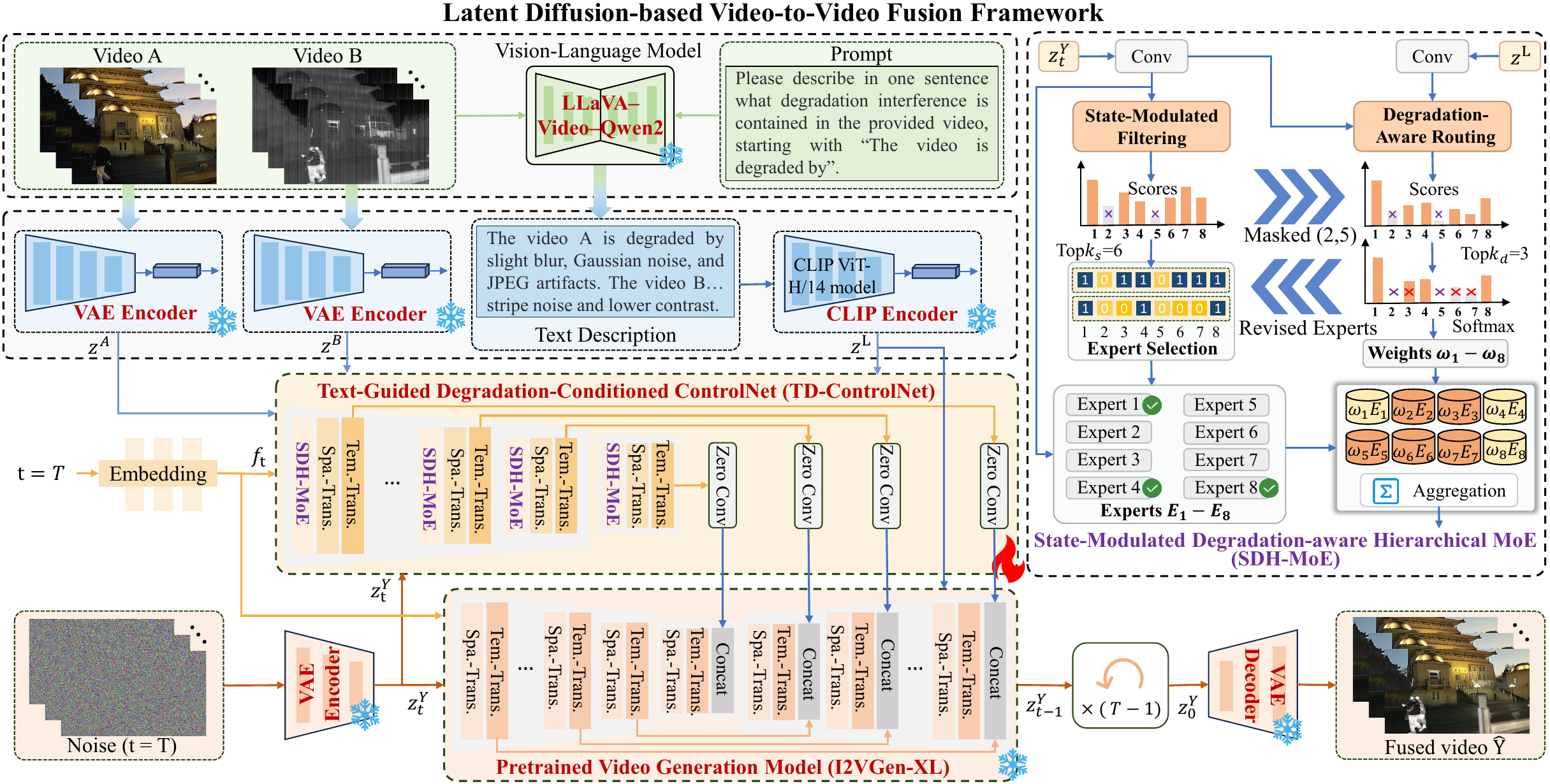}
    \vspace{-0.1in}
    \caption{Overall framework of the proposed V2VFusion. The pretrained I2VGen-XL backbone adopts a 3D U-Net with interleaved spatial and temporal transformer blocks to jointly model spatial appearance and temporal dynamics. TD-ControlNet follows the ControlNet paradigm by duplicating the encoder and middle block of the pretrained backbone and injecting the extracted conditional features into the frozen denoising U-Net through zero-initialized convolutions.}
    \label{network}
\end{figure*}

The overall framework is shown in Fig.~\ref{network}. Given degraded source videos $X_A, X_B \in \mathbb{R}^{F \times H \times W \times 3}$ ($F,H,W$ denoting frame numbers, height, and width), the goal is to generate a high-quality fused video $\hat{Y} \in \mathbb{R}^{F \times H \times W \times 3}$ that preserves complementary information while suppressing degradations.

 We first formulate video fusion as a conditional video-to-video generation problem by designing a latent diffusion-based video-to-video fusion framework. Built upon this framework, we introduce a text-guided degradation-conditioned ControlNet. It incorporates degraded video priors $z^A, z^B$ and a textual degradation prior $z^L$ into the diffusion process. By conditioning the generation on both visual cues and semantic guidance, the model can preserve structural information in source videos while explicitly reasoning about degradation characteristics, realizing degradation-robust video fusion.



Furthermore, to better handle compound degradations, we equip ControlNet with a state-modulated degradation-aware hierarchical mixture-of-experts module (SDH-MoE). Rather than relying on a single network to process diverse degradation patterns, this module adaptively routes intermediate features to specialized experts according to both diffusion states and degradation characteristics, enhancing the model’s capability to disentangle and suppress multiple degradation types.
\subsection{Latent Diffusion-based Video-to-Video Fusion Framework}
Rather than relying on frame-by-frame generation, we aim to directly model video fusion as a unified video generation problem. Inspired by recent Stable Diffusion-derived video generation models~\cite{rombach2022high, zhang2023i2vgen}, we formulate video fusion as a video-to-video conditioned generation task within a latent diffusion framework, enabling holistic end-to-end video fusion.

First, we employ a pretrained VAE encoder $\mathcal{E}$ \cite{blattmann2023stable}
to encode the ground-truth fused video $Y$ into a latent representation $z^Y = \mathcal{E}(Y)$, where $z^Y \in \mathbb{R}^{F \times h \times w \times C}$. Then, Gaussian noise is progressively added to $z^Y$ in forward process, modeled as a Markov chain with $T$ timesteps. The forward process defines a conditional distribution $q(\cdot)$, representing the noise corruption process, formulated as:
\begin{equation}
    q(z_t^Y | z_0^Y) = \mathcal{N}\left(z_t^Y; \sqrt{\hat{\alpha}_t} z_0^Y, \hat{\beta}_t \mathbf{I}\right),
\end{equation}
where $\mathbf{I}$ is an identity matrix. $\mathcal{N}(0;\textbf{I})$ is a standard normal distribution. $\beta_t$ is the variance schedule at timestep $t$. $\hat{\beta}_t = \prod_{i=1}^t \beta_i$, $\hat{\alpha}_t = \prod_{i=1}^t (1 - \beta_i)$. $z_t^Y$ is the latent noise at timestep $t$. As $t$ increases, $\beta_t$ gradually grows and $q(z_t^Y | z_0^Y)$ progressively approaches a standard normal distribution.


During the reverse denoising process, diffusion model starts from timestep $T$ and progressively generates a high-quality fused video from Gaussian noise through a denoising network $\epsilon_\theta$. To enable controllable and degradation-aware fusion, we further introduce two conditional priors: \textit{degraded video priors} and \textit{textual degradation priors}. Degraded videos provide scene-level visual priors to preserve spatial structures and motion dynamics of source observations. Textual descriptions provide high-level priors that explicitly indicate degradation types. By jointly leveraging complementary sources, the model can simultaneously reason about scene content and degradation characteristics.

Since priors originate from different modalities, representations are inherently heterogeneous. To enable effective conditioning within diffusion framework, both modalities are projected into latent space for feature alignment with diffusion variables.
\textit{For degraded video priors:}  $X_A,X_B$ are encoded by $\mathcal{E}$ to obtain latent representations $z^A$ and $z^B$.
\textit{For textual degradation priors:} a multi-modal vision–language model, LLaVA-Video-Qwen2~\cite{zhang2410video} analyzes the degraded source videos and generates degradation descriptions (\textit{e.g.}, \textit{“The visible video is degraded by $\cdots$. The infrared video is degraded by $\cdots$.”}) through an instructional prompt. Then, the textual descriptions are encoded by the pretrained CLIP text encoder (i.e., CLIP-ViT \cite{schuhmann2022laion}) to obtain a latent representation $z^L$.


To incorporate these priors into pretrained video diffusion model, we design conditional guidance via ControlNet framework~\cite{zhang2023adding}, which is primarily designed for image generation with single-modality control signals (e.g., edges or depth maps). When directly applied to video fusion, it faces two key limitations. First, it lacks mechanisms to jointly integrate heterogeneous conditional information from multiple modalities. Second, it does not explicitly model degradation characteristics, making it difficult to provide effective guidance under compound degradation conditions.

Thus, we develop a text-guided degradation-conditioned ControlNet (TD-ControlNet) $\pi_\theta$. It jointly incorporates degraded video priors and textual degradation semantics into the diffusion process. By bridging low-level visual degradation cues with high-level semantic descriptions, TD-ControlNet enables the diffusion model to adaptively regulate restoration and fusion (TD-ControlNet is detailed in Sec.~\ref{controlnet}). To preserve the capability of the pretrained video generation model while enabling effective conditional control, the outputs of TD-ControlNet are connected to the backbone through zero-initialized convolution layers. 
Finally, the reverse denoising process can be formulated as:
\begin{equation}
    p_\theta(z_{t-1}^{Y}|z_{t}^{Y}, z^A, z^B, z^L)= \mathcal{N}(z_{t-1}^{Y}; \mu_\theta(z_{t}^{Y},t,z^L,\pi_\theta(z_{t}^{Y}, z^A,z^B,t,z^L))),\sigma_t^2\textbf{I}),
\end{equation}
where $\mu_\theta$ is the mean and calculated by denoising network $\epsilon_\theta$. The variance $\sigma_t^2$ is a time-dependent constant. $p_\theta$ denotes a posterior distribution. Through Bayes' theorem, the true reverse posterior distribution can be derived as $p_\theta(z_{t-1}^{Y}|z_{t}^{Y}, z_0^Y)$. Thus, the mean $\hat{\mu}_t$ can be acquired in timestep $t$ by:
\begin{equation}
    \hat{\mu}_t(z_{t}^{Y},z_{0}^{Y}, z^A, z^B, z^L) = \frac{\sqrt{\hat{\alpha}_{t-1}}\beta_t}{1-\hat{\alpha}_t}z_0^Y + \frac{\sqrt{\alpha_{t}}(1-\hat{\alpha}_{t-1})}{1-\hat{\alpha}_t}z_t^Y.
\end{equation}
However, the clean latent variable $z_0^Y$ is unknown during inference. Therefore, diffusion models employ a denoising network $\epsilon_\theta$ to predict the noise component and estimate the clean sample:
\begin{equation}
    \hat{z}_0^Y = \frac{z_t^Y-\sqrt{1 - \hat{\alpha}_t}\epsilon_\theta(z_{t}^{Y},t,z^L,\pi_\theta(z_{t}^{Y}, z^A,z^B,t,z^L)))}{\sqrt{\hat{\alpha}_t}}.
\end{equation}
Substituting the estimated clean sample $\hat{z}_0^Y$ into the posterior mean formulation yields the practical parameterization used in diffusion models:
\begin{equation}
\footnotesize
    \mu_\theta(z_{t}^{Y},t, z^A, z^B, z^L) = \frac{z_t^Y-\frac{1-\alpha_t}{\sqrt{1-\hat{\alpha}_t}}\epsilon_\theta(z_{t}^{Y},t,z^L,\pi_\theta(z_{t}^{Y}, z^A,z^B,t,z^L)))}{\sqrt{\alpha_t}}, \quad \sigma_t^2 = \frac{1-\hat{\alpha}_{t-1}}{1-\alpha_t}\beta_t.
\end{equation}
The above formulation enables the model to iteratively predict the reverse transition and progressively remove noise, yielding the final latent representation $z_0^Y$. The generated fused video $\hat{Y}$ is then obtained by decoding $z_0^Y$ into the pixel space using the VAE decoder \cite{blattmann2023stable}.

\subsection{Text-Guided Degradation-Conditioned ControlNet} \label{controlnet}
To enable controllable and degradation-aware video fusion, we introduce a text-guided degradation-conditioned ControlNet (TD-ControlNet) that explicitly modulates the pretrained video diffusion model using both visual degradation cues and high-level textual degradation semantics. The key idea is to bridge low-level corrupted observations with semantic understanding, allowing the generation process to be guided by interpretable degradation priors rather than purely data-driven embeddings. 
Despite its effectiveness, representing diverse degradation types within a single unified representation space remains inherently limited.

Mixture-of-Experts (MoE) \cite{zhou2022mixture} provides a natural mechanism for modeling heterogeneous patterns by dynamically activating experts to handle compound degradations. However, effective routing typically relies on explicit conditional cues, while degradation information (e.g., blur or JPEG artifacts) is not always observable in latent diffusion representations. Moreover, the information structure of the latent variable $z_t^Y$ varies across diffusion timesteps. In early stages (large $t$), $z_t^Y$ is close to Gaussian noise and lacks discriminative structures, whereas in later stages (small $t$), textures gradually emerge, making degradations more identifiable. Consequently, applying a conventional MoE uniformly across the diffusion process may lead to unreliable routing due to insufficient degradation cues.

It motivates a more structured formulation of degradation modeling in diffusion models. Two factors should be jointly considered instead of the formulation as a unified task. First, the diffusion state determines recoverable information (e.g., noise or fine details) contained in the latent variable. Second, the degradation type determines the appropriate processing strategy. 
In essence, reliable degradation identification depends on an interpretable diffusion state, as the model can only infer degradation types once the currently observable information in latent space is determined.
\textit{From a probabilistic perspective}, expert selection can thus be decomposed into two components: a diffusion-state-dependent candidate expert set and degradation-aware routing within this set, formulated as:
\begin{equation}
P(E \mid z_t, f^c) = P(E \mid f^c, E \in \mathcal{S}(z_t)) \, P(E \in \mathcal{S}(z_t)|z_t),
\end{equation}
where $z_t$ denotes the latent variable at timestep $t$, $f^c$ represents the degradation-aware conditional representation, and $\mathcal{S}(z_t)$ denotes the candidate expert subset determined by the diffusion state.

\textbf{State-Modulated Degradation-aware Hierarchical Mixture-of-Experts (SDH-MoE)}\newline
Motivated by the above formulation, this module first performs a diffusion state-aware filtering step to determine which experts are compatible with the current denoising stage, and then conducts degradation-aware routing to select the most appropriate experts for handling the observed compound degradations. Specifically, in the state-aware filtering stage, given the latent feature $h_t$ derived from the diffusion state $z_t$ by in-layers, compatibility scores are predicted for each expert:
\begin{equation}
h_t = \text{in\_layers}(z_t), \quad s = g_s(h_t), \quad s = [s_1, s_2, \dots, s_K] \in \mathbb{R}^{K},
\end{equation}
where $g_s(\cdot)$ denotes the state-aware filtering network and $K$ is the total number of experts. Each score $s_i$ measures the compatibility between the $i$-th expert and the current diffusion state.
Based on these scores, a binary filtering mask $m \in \{0,1\}^{K}$ is constructed using a Top-$k$ operator. When $i \in \mathrm{TopK}(s, k_s)$, $m_i=1$. Otherwise, $m_i=0$, where $k_s$ denotes the number of candidate experts retained by the state-aware filtering stage. The resulting candidate expert subset can therefore be written as $\mathcal{S}(z_t) = \{ i \mid m_i = 1 \}$.
This diffusion-state-aware mask acts as a filtering mechanism that restricts the expert search space to those experts compatible with the current diffusion stage. As a result, unreliable expert activation can be avoided when the latent representation still contains insufficient structural information.

After obtaining the candidate expert subset $\mathcal{S}(z_t)$ through state-aware filtering, the degradation-aware router further performs fine-grained expert selection based on the degradation semantic representation $f^c$. Specifically, the router predicts degradation routing scores for all experts as:
\begin{equation}
f^c = \text{Context-Layers}(z^L),\quad r = g_d\big(h_t + f^c\big), \quad
r = [r_1, r_2, \dots, r_K] \in \mathbb{R}^{K},
\end{equation}
where $g_d(\cdot)$ is the degradation-aware gating network. Context-Layers (i.e., mean, SiLU and Linear) project the textual degradation representation into latent feature space, enabling alignment with $h_t$. To ensure that expert selection is restricted to the candidate subset determined by the diffusion state, the routing scores are filtered corresponding to $\mathcal{S}(z_t)$. When $i\in \mathcal{S}(z_t)$, $\tilde{r}_i=r_i$. Otherwise, $\tilde{r}_i=-\infty$. Based on it, the router performs a Top-$k$ selection within the candidate expert subset:
\begin{equation}
\mathcal{T}(z_t,f^c) = \mathrm{TopK}({\tilde{r}_i \mid i \in \mathcal{S}(z_t)}, k_d),
\end{equation}
where $k_d$ denotes the number of experts activated by the degradation-aware router. Subsequently, the routing weights are then computed using a softmax function over the selected experts:
\begin{equation}
\omega_i =
\frac{\exp(\tilde{r}_i)}
{\sum{j \in \mathcal{T}(z_t,f^c)} \exp(\tilde{r}_j)},
\quad i \in \mathcal{T}(z_t,f^c),
\end{equation}
where $\omega_i = 0$ for $i \notin \mathcal{T}(z_t,f^c)$. Finally, the SDH-MoE output is obtained by aggregating the outputs of the selected experts weighted by the routing coefficients:
\begin{equation}
h_t^{\text{moe}} =
\sum\nolimits_{i \in \mathcal{T}(z_t,f^c)} \omega_i E_i(h_t),
\end{equation}
where $E_i(\cdot)$ denotes the $i$-th expert network. Through this hierarchical routing design, SDH-MoE dynamically activates a compact subset of experts that are jointly compatible with the diffusion state and specialized for the underlying degradation, thereby improving routing reliability.

\subsection{Loss Functions}

To accelerate the diffusion sampling process, we adopt the progressive distillation paradigm to reduce the sampling cost. The optimization objective is defined as:
\begin{equation}
\mathcal{L}_{\text{Diff}} =
\mathbb{E}_{z_0^Y, \epsilon \sim \mathcal{N}(0, I), t}
\left[
\left\|
\sqrt{\hat{\alpha}_t}\epsilon - \sqrt{1-\hat{\alpha}_t}z_0^Y - \epsilon_\theta \left(z_t^Y, t, z^A, z^B, z^L \right)
\right\|_2^2
\right].
\end{equation}
To improve structural fidelity and suppress degradation artifacts, we introduce a frequency reconstruction loss. At timestep $t$, the generated fused video $\hat{Y}$ and ground-truth $Y$ are transformed into the frequency domain by Discrete Fourier Transform. A low-pass filter $\eta$ yields low/high frequency components as $\hat{f}_l = \mathcal{F}(\hat{Y}) \odot \eta$, $\hat{f}_h = \mathcal{F}(\hat{Y}) \odot (1 - \eta)$, and similarly $f_l$, $f_h$ for $Y$, where $\mathcal{F}(\cdot)$ is DFT. $\odot$ is Hadamard product. With a weighting control parameter $\psi$, the loss is defined as:
\begin{equation}
\mathcal{L}_{\text{Freq}}=(\frac{t}{T})^\psi
\| f_l - \hat{f}_l \|_2^2 + (1 - (\frac{t}{T})^\psi)\| f_h - \hat{f}_h \|_2^2.
\end{equation}
Finally, with $\lambda$ being a balancing coefficient, the overall loss function is $\mathcal{L} =\mathcal{L}_{\text{Diff}} + \lambda \mathcal{L}_{\text{Freq}}$.


\section{Experiment}
\textbf{Datasets and Implementation.}
We train and evaluate across three video fusion tasks: infrared–visible video fusion on M3SVD \cite{tang2025videofusion}, multi-exposure video fusion on YouTube-HDR \cite{zhao2025unified}, and multi-focus video fusion on DAVIS \cite{zhao2025unified}.
Various degradations are introduced during training, including blur, Gaussian noise, stripe noise, and JPEG compression artifacts. Based on high-quality source videos and existing video fusion methods~\cite{tang2025videofusion,gong2025temcoco,zhao2025unified}, we generate candidate fused videos and select high-quality results as ground truth. More details are provided in \textit{Supplementary Material}.

We train \textit{a unified fusion model} for multiple video fusion tasks. We use AdamW optimizer with initial learning rate of $5\times10^{-5}$, scheduled with 500 warmup steps followed by cosine annealing. Input videos are sampled at 30 FPS with a duration of 1 second. Hyperparameters are $\psi=2$, $\lambda=0.5$, $K=8$, $k_s=6$, $k_d=3$, $T=999$. The frequency mask $\eta$ is determined by the 80th-percentile magnitude threshold.
Experiments are conducted on four NVIDIA A800 GPUs with a batch size of 4 and trained for 15,000 iterations.

\begin{figure*}[!t]
    \centering    \includegraphics[width=1\linewidth]{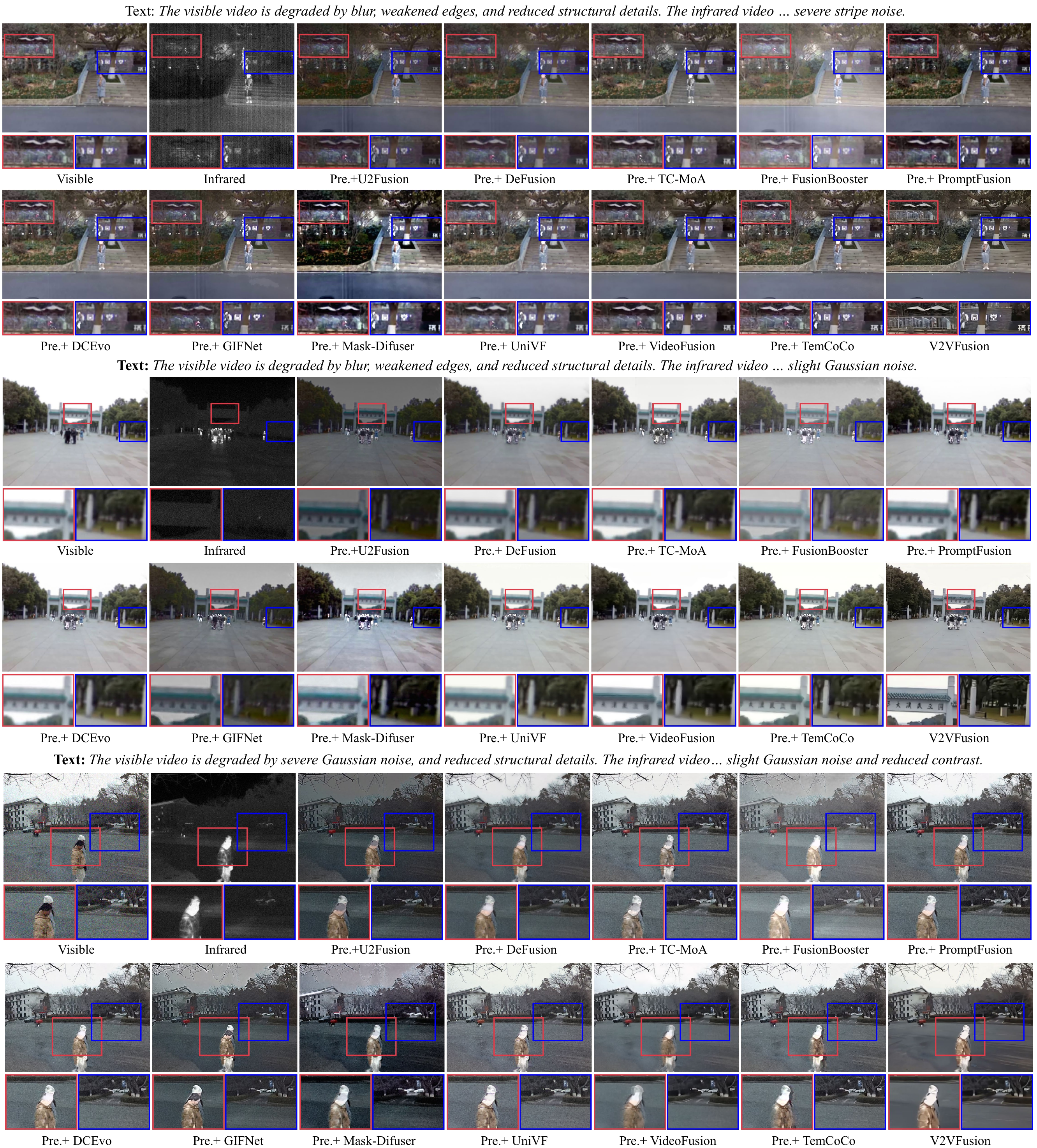}\\
    \vspace{-0.05in}
    \caption{Results on M3SVD infrared–visible dataset (Pre.: pre-enhancement \cite{chen2025spatiotemporal}, \cite{cai2024exploring}, \cite{pan2023deep}, \cite{liang2021swinir}).}
    \label{exper1}
\end{figure*}

\begin{figure*}[!t]
    \centering    \includegraphics[width=1\linewidth]{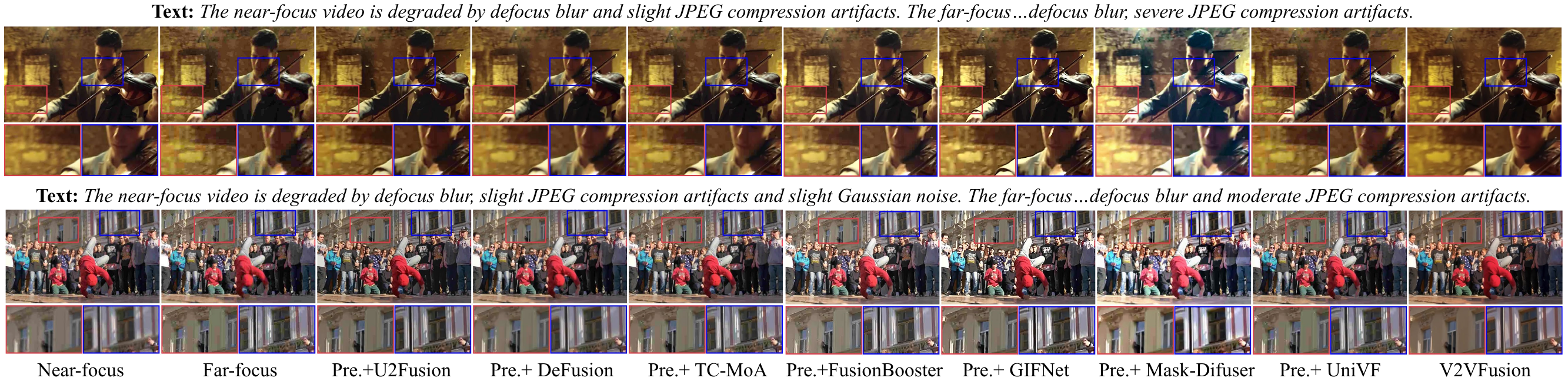}\\
    \vspace{-0.05in}
    \caption{Results on DAVIS multi-focus dataset (Pre.: pre-enhancement \cite{chen2025spatiotemporal, cai2024exploring, pan2023deep, liang2021swinir}).}
    \label{exper2}
\end{figure*}
\begin{figure*}[!t]
    \centering    \includegraphics[width=1\linewidth]{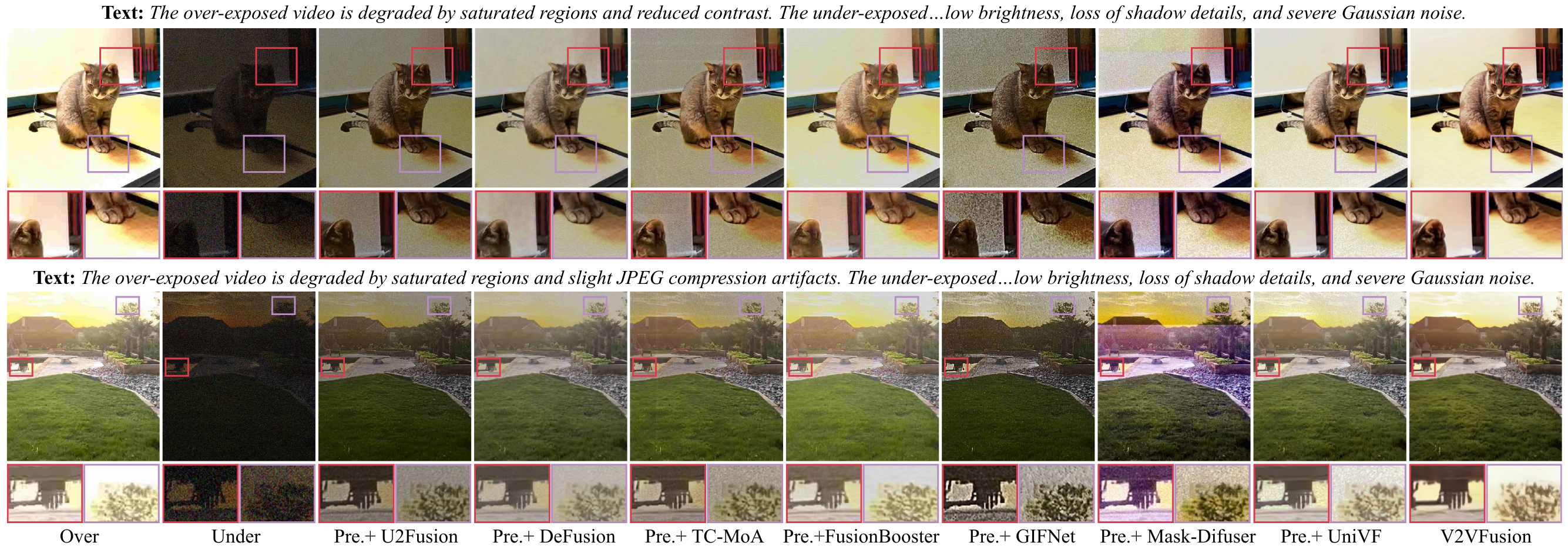}\\
    \vspace{-0.05in}
    \caption{Results on YouTube. multi-exposure dataset (Pre.: pre-enhancement \cite{chen2025spatiotemporal}, \cite{cai2024exploring}, \cite{pan2023deep}, \cite{liang2021swinir}).}
    \label{exper3}
\end{figure*}

\textbf{Results and Analysis.} We compare V2VFusion with several SOTA image fusion methods (U2Fusion~\cite{xu2020u2fusion}, DeFusion~\cite{liang2022fusion}, TC-MoA~\cite{zhu2024task}, FusionBooster~\cite{cheng2025fusionbooster}, PromptFusion~\cite{liu2024promptfusion}, DCEvo~\cite{liu2025dcevo}, GIFNet~\cite{cheng2025one}, Mask-Difuser~\cite{tang2025mask}) and video fusion approaches (VideoFusion~\cite{tang2025videofusion}, TemCoCo~\cite{gong2025temcoco}, and UniVF~\cite{zhao2025unified}). 
For fair comparison under degradation conditions, we employ several SOTA pre-enhancement methods, including visible denoising \cite{chen2025spatiotemporal}, infrared denoising \cite{cai2024exploring}, deblurring \cite{pan2023deep}, and JPEG artifact removal \cite{liang2021swinir}, which are retrained under default experimental settings.

\textit{\textbf{Qualitative Comparison.}} Results across multiple video fusion tasks and diverse degradation conditions are shown in Figs.~\ref{exper1}-\ref{exper3}. More qualitative comparisons and additional evaluations on challenging temporal scenarios, including fast motion and long video sequences, are provided in the \textit{Supplementary Material}. We analyze the results from three aspects.


\textit{1) Fusion Quality: our method achieves clearer structures and richer textures.}  
Existing methods often produce flickering and texture distortions due to limited temporal consistency, whereas our video‑level generation framework achieves more stable coherence while preserving fine spatial details.

\textit{2) Degradation Suppression: our method is more effective at mitigating compound degradations.} 
Although some existing fusion pipelines combine pre-enhancement modules with fusion networks, residual degradations still remain due to the limited synergy between enhancement and fusion. In contrast, our unified text-guided multi-degradation-aware framework jointly performs restoration and fusion, thereby suppressing various degradations more effectively.

\begin{figure*}[!t]
    \centering    \includegraphics[width=1\linewidth]{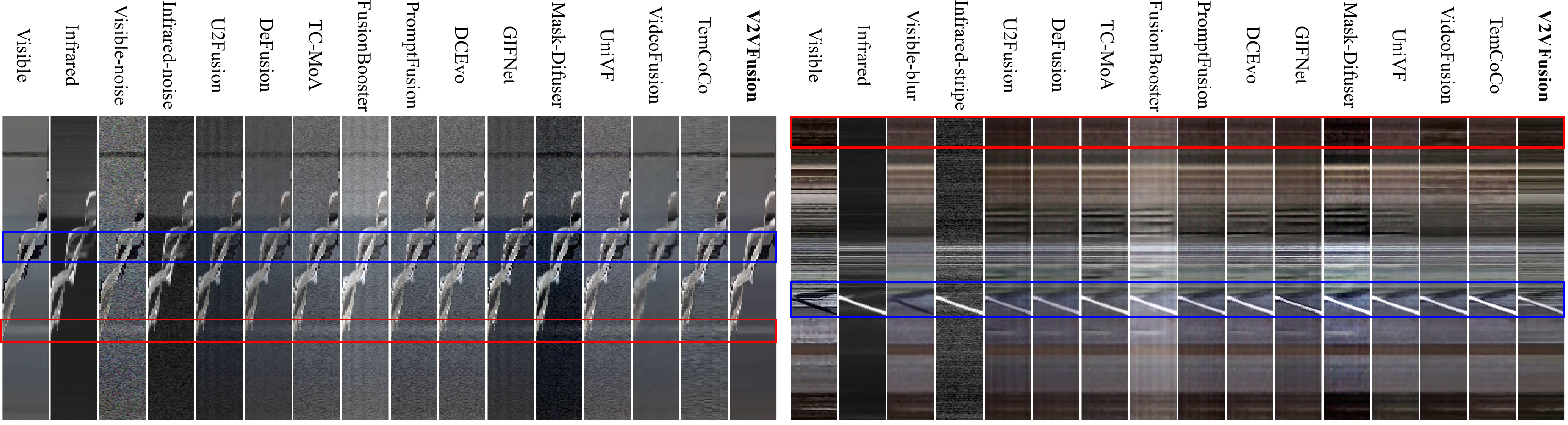}\\
    \vspace{-0.05in}
    \caption{Comparison of temporal profiles (competing methods adopt pre-enhancement).}
    \label{exper-tem1}
\end{figure*}

\begin{figure*}[!t]
    \centering    \includegraphics[width=1\linewidth]{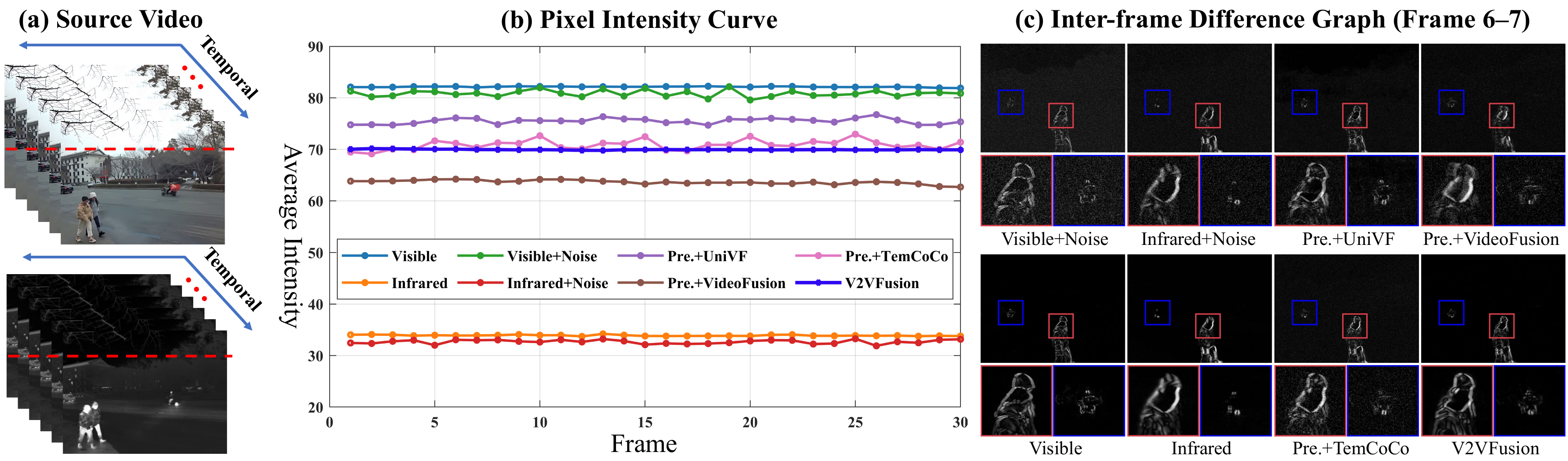}\\
    \vspace{-0.05in}
    \caption{Comparison of pixel intensity curve and inter-frame difference (Pre.: pre-enhancement).}
    \label{exper-tem2}
\end{figure*}

\textit{3) Temporal Consistency: our method demonstrates superior temporal consistency.} \textit{i) Temporal profiles:} In Fig.~\ref{exper-tem1}, competitors exhibit noticeable temporal noise and structural discontinuities with severe fluctuations, whereas our method yields the most stable temporal structures with substantially reduced flickering and degradation artifacts. \textit{ii) Pixel intensity curves:} In Fig.~\ref{exper-tem2}(b), we assess temporal smoothness via pixel intensity variations across frames. Our method maintains consistent intensity transitions, while competitors display irregular fluctuations indicative of inferior temporal stability. \textit{iii) Inter-frame differences:} In Fig.~\ref{exper-tem2}(c), we compute consecutive frame differences. Our method produces more motion-aligned difference patterns, whereas competitors exhibit motion artifacts and residual degradation, reflecting weaker temporal coherence.


\begin{table*}[!t]
\centering
\caption{Results on three video fusion tasks (P.: pre-enhancement; \textcolor{red}{\textbf{red}}: optimal; \textcolor{blue}{\textbf{blue}}: suboptimal).}
\label{table_ir_visible}
\resizebox{1\linewidth}{!}{
\renewcommand\arraystretch{1.15}
\begin{tabular}{l@{\,~}|c@{\,~~~}c@{\,~~~}c@{\,~~~}c@{\,~~~}c@{\,~~~}c@{\,~~~}c@{\,~~~}c@{\,~~~}c@{\,~~~}c@{\,~~~}c@{\,~~~}c@{\,~}}
\toprule
\multirow{2}{*}{\textbf{Metric}}&
\multicolumn{12}{c}{\textbf{Infrared-Visible (M3SVD)}} \\
\cmidrule(lr){2-13}&
P.+U2Fus. &
P.+DeFus. &
P.+TC. &
P.+FusionB. &
P.+Prompt. &
P.+DCEvo &
P.+GIF. &
P.+Mask. &
P.+VideoFus. &
P.+TemCoCo &
P.+UniVF &
\textbf{V2VFusion} \\
\midrule
\textbf{MI$\uparrow$} &
1.982 & 2.224 & 2.002 & 2.118 & 2.210 & 2.457 & 2.433 & 2.190 & 2.523 & \BH{2.661} & 2.31 & \RH{2.684} \\
\textbf{Qabf$\uparrow$} &
0.393 & 0.486 & 0.536 & 0.510 & 0.592 & \RH{0.671} & \BH{0.653} & 0.536 & 0.622 & 0.634 & 0.626 & 0.607 \\
\textbf{tSSIM$\uparrow$} &
0.942 & 0.914 & 0.866 & 0.876 & 0.858 & 0.852 & 0.864 & 0.850 & \BH{0.972} & 0.953 & 0.885 & \RH{0.983} \\
\textbf{tF$\downarrow$} &
1.402 & 1.496 & 2.285 & 2.288 & 2.268 & 2.331 & 2.502 & 2.830 & \BH{0.792} & 1.404 & 2.256 & \RH{0.514} \\
\textbf{tLP$\downarrow$} &
0.352 & 0.535 & 0.536 & 0.578 & 0.599 & 0.580 & 0.610 & 0.614 & \BH{0.155} & 0.354 & 0.493 & \RH{0.110} \\
\textbf{BiSWE$\downarrow$} &
3.502 & 3.155 & 4.456 & 4.608 & 4.559 & 4.660 & 5.000 & 5.591 & \BH{1.592} & 3.503 & 4.025 & \RH{0.881} \\
\bottomrule
\end{tabular}}
\resizebox{1\linewidth}{!}{
\renewcommand\arraystretch{1.15}
\begin{tabular}{l@{\,~}|c@{\,~~}c@{\,~~}c@{\,~~}c@{\,~~}c@{\,~~}c@{\,~~}c@{\,~~}c@{\,~}|c@{\,~~}c@{\,~~}c@{\,~~}c@{\,~~}c@{\,~~}c@{\,~~}c@{\,~~}c@{\,~}}
\toprule
\multirow{2}{*}{\textbf{Metric}} &
\multicolumn{8}{c|}{\textbf{Multi-Exposure (YouTube-HDR)}} &
\multicolumn{8}{c}{\textbf{Multi-Focus (DAVIS)}} \\
\cmidrule(lr){2-9} \cmidrule(lr){10-17}
& P.+U2Fus. & P.+DeFus. & P.+TC. & P.+FusionB. & P.+GIF. & P.+Mask. & P.+UniVF & \textbf{V2VFusion}
& P.+U2Fus. & P.+DeFus. & P.+TC. & P.+FusionB. & P.+GIF. & P.+Mask. & P.+UniVF & \textbf{V2VFusion} \\
\midrule
\textbf{MI$\uparrow$}
& 2.331 & 2.644 & 3.912 & 2.986 & 4.197 & 3.048 & \BH{4.318} & \RH{4.495}
& 3.563 & 4.783 & 5.346& 5.662 & \BH{6.343} & 5.117 & 6.023 & \RH{6.489} \\
\textbf{Qabf$\uparrow$}
& 0.369 & 0.577 & 0.544 & 0.613 & \BH{0.692} & 0.435 & 0.686 & \RH{0.739}
& 0.615 & 0.553 & 0.634 & \BH{0.726} & 0.712 & 0.520 & 0.701 & \RH{0.778} \\
\textbf{tSSIM$\uparrow$}
& 0.592 & 0.683 & 0.542 & 0.681 & \BH{0.444} & 0.495 & 0.626 & \RH{0.767}
& 0.598 & \BH{0.658} & 0.585 & 0.553 & 0.542 & 0.575 & 0.577 & \RH{0.669} \\
\textbf{tF$\downarrow$}
& \BH{4.157} & 4.856 & 5.204 & 4.163 & 7.852 & 7.193 & 5.151 & \RH{3.882}
& 5.823 & 5.905 & 5.807 & 6.819 & 8.147 & 7.905 & \BH{4.871} & \RH{4.710} \\
\textbf{tLP$\downarrow$}
& 0.142 & 0.133 & 0.152 & 0.131 & 0.185 & 0.177 & \BH{0.127} & \RH{0.075}
& 0.186 & 0.207 & 0.193 & 0.192 & 0.191 & 0.205 & \BH{0.177} & \RH{0.116} \\
\textbf{BiSWE$\downarrow$}
& 4.584 & 4.086 & 5.909 & 4.553 & 9.902 & 8.984 & \BH{4.578} & \RH{2.566}
& 5.354 & \BH{4.595} & 5.703& 6.431 & 7.243 & 7.948 & 6.092 & \RH{3.231} \\
\bottomrule
\end{tabular}}
\end{table*}

\textit{\textbf{Quantitative Comparison.}} We evaluate with spatial quality metrics (MI, Qabf) to assess information fidelity and temporal consistency metrics (tSSIM, tF, tLP, BiSWE) to measure the stability of inter-frame representations and temporal fluctuations. As reported in Tab.~\ref{table_ir_visible}, our method consistently achieves superior performance across all benchmarks. Improvements in MI and Qabf indicate that the proposed framework can better preserve complementary information and structural details in source videos. Advantages in tSSIM, tF, tLP, and BiSWE show that our fused videos exhibit more stable temporal structures with fewer frame-to-frame inconsistencies and reduced flickering artifacts, validating the effectiveness of the proposed video-to-video generation paradigm (see \textit{Supplementary Material} for the no-reference quality evaluation results).

\begin{table*}[!t]
\centering
\caption{Quantitative results of ablation study.}
\label{table_ablation}
\resizebox{1\linewidth}{!}{
\renewcommand\arraystretch{1.15}
\begin{tabular}{l@{\,~~}|c@{\,~~}c@{\,~~}c@{\,~~}c@{\,~~}c@{\,~~}c@{\,~}|c@{\,~~}c@{\,~~}c@{\,~~}c@{\,~~}c@{\,~~}c@{\,~}|
c@{\,~~}c@{\,~~}c@{\,~~}c@{\,~~}c@{\,~~}c@{\,~}}
\toprule
\multirow{2}{*}{\textbf{Index}} &
\multicolumn{6}{c|}{\textbf{Infrared-Visible (M3SVD)}} &
\multicolumn{6}{c|}{\textbf{Multi-Exposure (YouTube-HDR)}} &
\multicolumn{6}{c}{\textbf{Multi-Focus (DAVIS)}}\\
\cmidrule(lr){2-7} \cmidrule(lr){8-13} \cmidrule(lr){14-19}
&\textbf{MI$\uparrow$}&\textbf{Qabf$\uparrow$}& \textbf{tSSIM$\uparrow$}&\textbf{tF$\downarrow$}& \textbf{tLP$\downarrow$}&\textbf{BiSWE$\downarrow$}& \textbf{MI$\uparrow$}&\textbf{Qabf$\uparrow$}& \textbf{tSSIM$\uparrow$}&\textbf{tF$\downarrow$}&
\textbf{tLP$\downarrow$}&\textbf{BiSWE$\downarrow$}& 
\textbf{MI$\uparrow$}&\textbf{Qabf$\uparrow$}& \textbf{tSSIM$\uparrow$}&\textbf{tF$\downarrow$}&
\textbf{tLP$\downarrow$}&\textbf{BiSWE$\downarrow$}\\
\midrule
w/o $\mathcal{L}_{\text{Freq}}$ &2.45 &0.55 &0.95 &0.57 & 0.15&0.90 &4.30 & 0.69 &0.72 &3.97 &0.08 &2.59 &6.39 &0.72 &0.61 &4.80 &0.13 &\RH{3.22} \\
w/o Text & 2.38&0.52 &0.92 &0.60 &0.14 &0.93 &4.27  &0.61 &0.69 &4.01 &0.10 &2.63 &6.31 &0.67 &0.59 &4.82 &0.15 &3.31 \\
w/o SDH-MoE & 2.53&0.56 &0.97 & 0.58& 0.12 &0.89 &4.38 &0.67 &0.70 & 3.90& 0.11& 2.58&6.29 &0.68 &0.59 &4.78 &0.16 &3.27  \\
w/o State & 2.59&0.58 &\RH{0.99} & 0.54& 0.13& 0.90& 4.44 & 0.71& 0.73& 3.92& 0.09& 2.59&6.43 &0.74 &0.62 &4.73 &0.15 & 3.24\\
\textbf{V2VFusion} & \RH{2.68}&\RH{0.60}&0.98 & \RH{0.51}&\RH{0.11} & \RH{0.88}& \RH{4.49} & \RH{0.73}& \RH{0.76}& \RH{3.88}&\RH{0.07} &\RH{2.56} &\RH{6.48} &\RH{0.77} & \RH{0.66}&\RH{4.71} &\RH{0.11} &3.23 \\
\bottomrule
\end{tabular}}
\end{table*}

\begin{figure*}[!t]
    \centering    \includegraphics[width=1\linewidth]{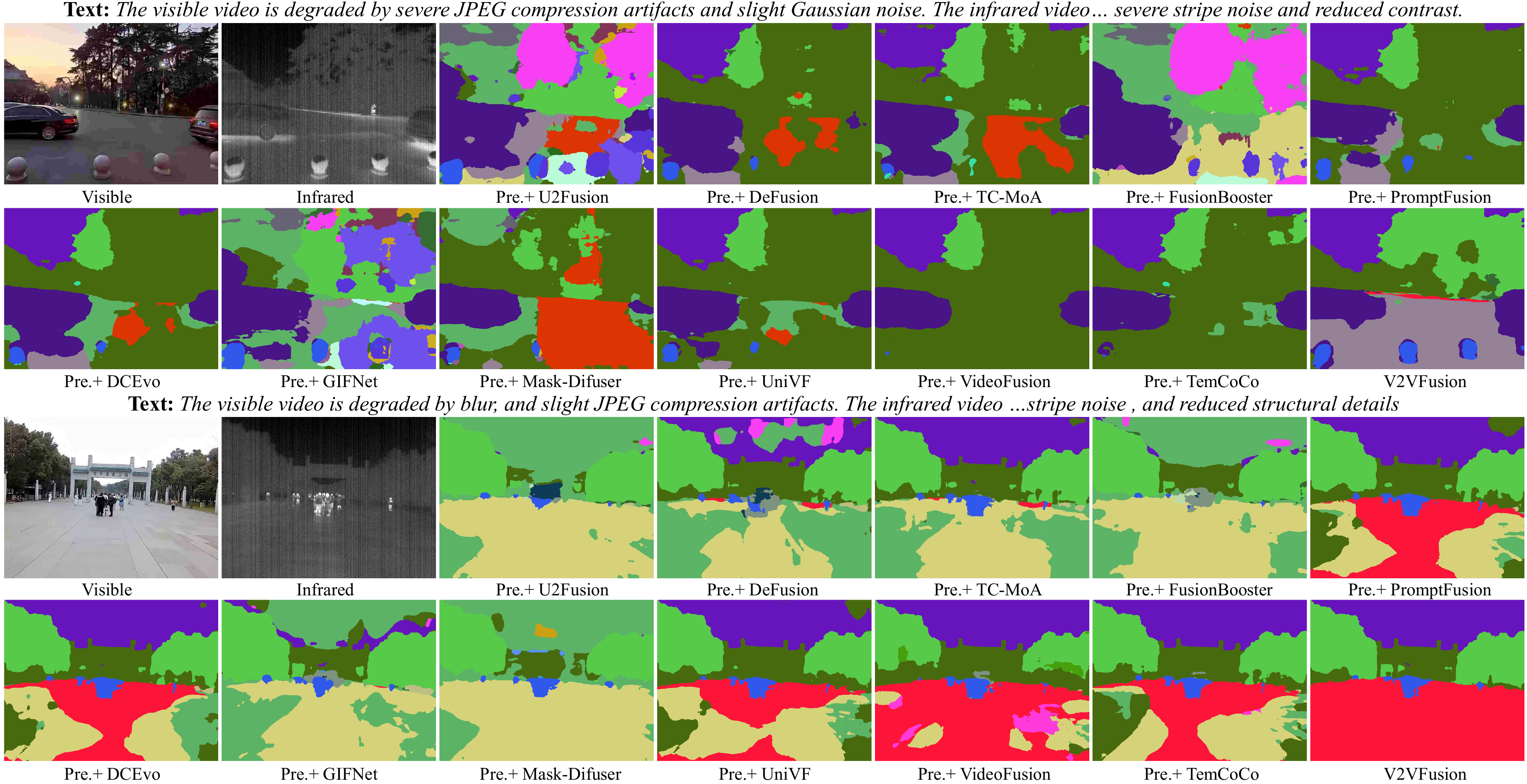}\\
    \vspace{-0.05in}
    \caption{Qualitative comparison results of semantic segmentation (Pre.: pre-enhancement).}
    \label{seg}
\end{figure*}
\textbf{Ablation Study.} We conduct ablation studies by progressively removing different modules: 
\textbf{I}: removing the frequency reconstruction loss (w/o $\mathcal{L}_{\text{Freq}}$); 
\textbf{II}: removing the textual degradation description (w/o Text); 
\textbf{III}: removing the proposed SDH-MoE module (w/o SDH-MoE); 
\textbf{IV}: removing the state-modulated mechanism within SDH-MoE (w/o State). As shown in Tab.~\ref{table_ablation}, removing individual components generally leads to a decline in overall performance across different tasks and evaluation metrics, demonstrating the contribution of each component to the effectiveness of the proposed framework (Qualitative comparisons are further provided in \textit{Supplementary Material}).

\textbf{Semantic Verification on High-Level Task.} 
To further evaluate the effectiveness of our method in preserving and enhancing high-level semantic information, we conduct experiments on semantic segmentation using SegFormer \cite{xie2021segformer}.  As shown in Fig.~\ref{seg}, our method produces more semantically coherent and structurally consistent predictions, especially in regions with severe degradations. This indicates that the proposed fusion framework not only improves low-level visual quality but also effectively preserves discriminative features required for high-level semantic understanding.

\section{Conclusion}
This paper presents V2VFusion, a unified video-to-video fusion framework that generates temporally coherent results by conditional generation in latent diffusion space. A text-guided degradation-conditioned ControlNet and a state-modulated degradation-aware MoE respectively bridge semantic control and handle compound degradations. Experiments demonstrate its effectiveness.

\begin{ack}
This work was supported by the National Natural Science Foundation of China (NSFC) under Grant 62502085 and U21B2027, the Natural Science Foundation of Jiangsu Province of China under Grant BK20241280, and the Start-up Research Fund of Southeast University under Grant RF1028624061 and RF1028623006.
\end{ack}
\bibliography{ref}

@article{li2018densefuse,
  title={DenseFuse: A fusion approach to infrared and visible images},
  author={Li, Hui and Wu, Xiao-Jun},
  journal={IEEE Transactions on Image Processing},
  volume={28},
  number={5},
  pages={2614--2623},
  year={2018},
  publisher={IEEE}
}

@article{li2020nestfuse,
  title={NestFuse: An infrared and visible image fusion architecture based on nest connection and spatial/channel attention models},
  author={Li, Hui and Wu, Xiao-Jun and Durrani, Tariq},
  journal={IEEE Transactions on Instrumentation and Measurement},
  volume={69},
  number={12},
  pages={9645--9656},
  year={2020},
  publisher={IEEE}
}

@inproceedings{xu2020fusiondn,
  title={Fusiondn: A unified densely connected network for image fusion},
  author={Xu, Han and Ma, Jiayi and Le, Zhuliang and Jiang, Junjun and Guo, Xiaojie},
  booktitle={Proceedings of the AAAI Conference on Artificial Intelligence},
  volume={34},
  number={07},
  pages={12484--12491},
  year={2020}
}

@article{xu2020u2fusion,
  title={U2Fusion: A unified unsupervised image fusion network},
  author={Xu, Han and Ma, Jiayi and Jiang, Junjun and Guo, Xiaojie and Ling, Haibin},
  journal={IEEE Transactions on Pattern Analysis and Machine Intelligence},
  volume={44},
  number={1},
  pages={502--518},
  year={2020},
  publisher={IEEE}
}

@inproceedings{cao2023multi,
  title={Multi-modal gated mixture of local-to-global experts for dynamic image fusion},
  author={Cao, Bing and Sun, Yiming and Zhu, Pengfei and Hu, Qinghua},
  booktitle={Proceedings of the IEEE/CVF International Conference on Computer Vision},
  pages={23555--23564},
  year={2023}
}

@article{ma2019fusiongan,
  title={FusionGAN: A generative adversarial network for infrared and visible image fusion},
  author={Ma, Jiayi and Yu, Wei and Liang, Pengwei and Li, Chang and Jiang, Junjun},
  journal={Information fusion},
  volume={48},
  pages={11--26},
  year={2019},
  publisher={Elsevier}
}

@article{ma2020ddcgan,
  title={DDcGAN: A dual-discriminator conditional generative adversarial network for multi-resolution image fusion},
  author={Ma, Jiayi and Xu, Han and Jiang, Junjun and Mei, Xiaoguang and Zhang, Xiao-Ping},
  journal={IEEE Transactions on Image Processing},
  volume={29},
  pages={4980--4995},
  year={2020},
  publisher={IEEE}
}

@article{ma2022swinfusion,
  title={SwinFusion: Cross-domain long-range learning for general image fusion via swin transformer},
  author={Ma, Jiayi and Tang, Linfeng and Fan, Fan and Huang, Jun and Mei, Xiaoguang and Ma, Yong},
  journal={IEEE/CAA Journal of Automatica Sinica},
  volume={9},
  number={7},
  pages={1200--1217},
  year={2022},
  publisher={IEEE}
}

@article{rao2023tgfuse,
  title={TGFuse: An infrared and visible image fusion approach based on transformer and generative adversarial network},
  author={Rao, Dongyu and Xu, Tianyang and Wu, Xiao-Jun},
  journal={IEEE Transactions on Image Processing},
  year={2023},
  publisher={IEEE}
}

@article{yi2024diff,
  title={Diff-IF: Multi-modality image fusion via diffusion model with fusion knowledge prior},
  author={Yi, Xunpeng and Tang, Linfeng and Zhang, Hao and Xu, Han and Ma, Jiayi},
  journal={Information Fusion},
  volume={110},
  pages={102450},
  year={2024},
  publisher={Elsevier}
}

@inproceedings{tang2024drmf,
  title={DRMF: Degradation-robust multi-modal image fusion via composable diffusion prior},
  author={Tang, Linfeng and Deng, Yuxin and Yi, Xunpeng and Yan, Qinglong and Yuan, Yixuan and Ma, Jiayi},
  booktitle={Proceedings of the ACM International Conference on Multimedia},
  pages={8546--8555},
  year={2024}
}

@article{yue2023dif,
  title={Dif-fusion: Toward high color fidelity in infrared and visible image fusion with diffusion models},
  author={Yue, Jun and Fang, Leyuan and Xia, Shaobo and Deng, Yue and Ma, Jiayi},
  journal={IEEE Transactions on Image Processing},
  volume={32},
  pages={5705--5720},
  year={2023},
  publisher={IEEE}
}

@inproceedings{yi2024text,
  title={Text-if: Leveraging semantic text guidance for degradation-aware and interactive image fusion},
  author={Yi, Xunpeng and Xu, Han and Zhang, Hao and Tang, Linfeng and Ma, Jiayi},
  booktitle={Proceedings of the IEEE/CVF Conference on Computer Vision and Pattern Recognition},
  pages={27026--27035},
  year={2024}
}

@inproceedings{liang2022fusion,
  title={Fusion from decomposition: A self-supervised decomposition approach for image fusion},
  author={Liang, Pengwei and Jiang, Junjun and Liu, Xianming and Ma, Jiayi},
  booktitle={Proceedings of the European Conference on Computer Vision},
  pages={719--735},
  year={2022},
  organization={Springer}
}

@inproceedings{liu2022target,
  title={Target-aware dual adversarial learning and a multi-scenario multi-modality benchmark to fuse infrared and visible for object detection},
  author={Liu, Jinyuan and Fan, Xin and Huang, Zhanbo and Wu, Guanyao and Liu, Risheng and Zhong, Wei and Luo, Zhongxuan},
  booktitle={Proceedings of the IEEE/CVF Conference on Computer Vision and Pattern Recognition},
  pages={5802--5811},
  year={2022}
}

@article{zhang2024text,
  title={Text-difuse: An interactive multi-modal image fusion framework based on text-modulated diffusion model},
  author={Zhang, Hao and Cao, Lei and Ma, Jaiyi},
  journal={In the Thirty-eighth Annual Conference on Advances in Neural Information Processing Systems},
  volume={37},
  pages={39552--39572},
  year={2024}
}

@inproceedings{gong2025temcoco,
  title={Temcoco: Temporally consistent multi-modal video fusion with visual-semantic collaboration},
  author={Gong, Meiqi and Zhang, Hao and Yi, Xunpeng and Tang, Linfeng and Ma, Jiayi},
  booktitle={Proceedings of the IEEE/CVF International Conference on Computer Vision},
  pages={14326--14335},
  year={2025}
}

@article{tang2025videofusion,
  title={VideoFusion: A spatio-temporal collaborative network for multi-modal video fusion and restoration},
  author={Tang, Linfeng and Wang, Yeda and Gong, Meiqi and Li, Zizhuo and Deng, Yuxin and Yi, Xunpeng and Li, Chunyu and Xu, Han and Zhang, Hao and Ma, Jiayi},
  journal={arXiv preprint arXiv:2503.23359},
  year={2025}
}

@article{zhao2025unified,
  title={A unified solution to video fusion: From multi-frame learning to benchmarking},
  author={Zhao, Zixiang and Bai, Haowen and Ke, Bingxin and Cui, Yukun and Deng, Lilun and Zhang, Yulun and Zhang, Kai and Schindler, Konrad},
  journal={arXiv preprint arXiv:2505.19858},
  year={2025}
}

@inproceedings{xu2025deno,
  title={Deno-IF: Unsupervised Noisy Visible and Infrared Image Fusion Method},
  author={Xu, Han and Li, Yuyang and Deng, Yunfei and Ma, Jiayi and Liu, Guangcan},
  booktitle={the Thirty-ninth Annual Conference on Neural Information Processing Systems},
  year={2025}
}

@article{xu2025urfusion,
  title={Urfusion: Unsupervised unified degradation-robust image fusion network},
  author={Xu, Han and Yi, Xunpeng and Lu, Chen and Liu, Guangcan and Ma, Jiayi},
  journal={IEEE Transactions on Image Processing},
  year={2025},
  publisher={IEEE}
}

@inproceedings{cao2025mmaif,
  title={Mmaif: Multi-task and multi-degradation all-in-one for image fusion with language guidance},
  author={Cao, Zihan and Zhong, Yu and Wang, Ziqi and Deng, Liang-Jian},
  booktitle={Proceedings of the IEEE/CVF International Conference on Computer Vision},
  pages={11744--11754},
  year={2025}
}

@article{li2025towards,
  title={Towards Unified Semantic and Controllable Image Fusion: A Diffusion Transformer Approach},
  author={Li, Jiayang and Jiang, Chengjie and Jiang, Junjun and Liang, Pengwei and Ma, Jiayi and Nie, Liqiang},
  journal={IEEE Transactions on Pattern Analysis and Machine Intelligence},
  year={2025},
  publisher={IEEE}
}

@article{zhang2023i2vgen,
  title={I2vgen-xl: High-quality image-to-video synthesis via cascaded diffusion models},
  author={Zhang, Shiwei and Wang, Jiayu and Zhang, Yingya and Zhao, Kang and Yuan, Hangjie and Qin, Zhiwu and Wang, Xiang and Zhao, Deli and Zhou, Jingren},
  journal={arXiv preprint arXiv:2311.04145},
  year={2023}
}

@inproceedings{rombach2022high,
  title={High-resolution image synthesis with latent diffusion models},
  author={Rombach, Robin and Blattmann, Andreas and Lorenz, Dominik and Esser, Patrick and Ommer, Bj{\"o}rn},
  booktitle={Proceedings of the IEEE/CVF Conference on Computer Vision and Pattern Recognition},
  pages={10684--10695},
  year={2022}
}

@inproceedings{zhang2023adding,
  title={Adding conditional control to text-to-image diffusion models},
  author={Zhang, Lvmin and Rao, Anyi and Agrawala, Maneesh},
  booktitle={Proceedings of the IEEE/CVF International Conference on Computer Vision},
  pages={3836--3847},
  year={2023}
}

@article{blattmann2023stable,
  title={Stable video diffusion: Scaling latent video diffusion models to large datasets},
  author={Blattmann, Andreas and Dockhorn, Tim and Kulal, Sumith and Mendelevitch, Daniel and Kilian, Maciej and Lorenz, Dominik and Levi, Yam and English, Zion and Voleti, Vikram and Letts, Adam and others},
  journal={arXiv preprint arXiv:2311.15127},
  year={2023}
}

@article{zhang2410video,
  title={Video instruction tuning with synthetic data, 2024},
  author={Zhang, Yuanhan and Wu, Jinming and Li, Wei and Li, Bo and Ma, Zejun and Liu, Ziwei and Li, Chunyuan},
  journal={URL https://arxiv. org/abs/2410.02713},
  volume={17}
}

@article{schuhmann2022laion,
  title={Laion-5b: An open large-scale dataset for training next generation image-text models},
  author={Schuhmann, Christoph and Beaumont, Romain and Vencu, Richard and Gordon, Cade and Wightman, Ross and Cherti, Mehdi and Coombes, Theo and Katta, Aarush and Mullis, Clayton and Wortsman, Mitchell and others},
  journal={In the Thirty-sixth Annual Conference on Advances in Neural Information Processing Systems},
  volume={35},
  pages={25278--25294},
  year={2022}
}

@article{muller2009cognitively,
  title={Cognitively-engineered multisensor image fusion for military applications},
  author={Muller, Amanda C and Narayanan, Sundaram},
  journal={Information Fusion},
  volume={10},
  number={2},
  pages={137--149},
  year={2009},
  publisher={Elsevier}
}

@article{zhou2022mixture,
  title={Mixture-of-experts with expert choice routing},
  author={Zhou, Yanqi and Lei, Tao and Liu, Hanxiao and Du, Nan and Huang, Yanping and Zhao, Vincent and Dai, Andrew M and Le, Quoc V and Laudon, James and others},
  journal={Advances in Neural Information Processing Systems},
  volume={35},
  pages={7103--7114},
  year={2022}
}

@article{ding2026cmvf,
  title={CMVF: Cross-Modal Unregistered Video Fusion via Spatio-Temporal Consistency},
  author={Ding, Jianfeng and Zhang, Hao and Wang, Zhongyuan and Xiao, Jinsheng and Tian, Xin and Han, Zhen and Ma, Jiayi},
  journal={Information Fusion},
  pages={104212},
  year={2026},
  publisher={Elsevier}
}

@inproceedings{zhu2024task,
  title={Task-customized mixture of adapters for general image fusion},
  author={Zhu, Pengfei and Sun, Yang and Cao, Bing and Hu, Qinghua},
  booktitle={Proceedings of the IEEE/CVF Conference on Computer Vision and Pattern Recognition},
  pages={7099--7108},
  year={2024}
}

@article{cheng2025fusionbooster,
  title={Fusionbooster: A unified image fusion boosting paradigm},
  author={Cheng, Chunyang and Xu, Tianyang and Wu, Xiao-Jun and Li, Hui and Li, Xi and Kittler, Josef},
  journal={International Journal of Computer Vision},
  volume={133},
  number={5},
  pages={3041--3058},
  year={2025},
  publisher={Springer}
}

@article{liu2024promptfusion,
  title={PromptFusion: Harmonized semantic prompt learning for infrared and visible image fusion},
  author={Liu, Jinyuan and Li, Xingyuan and Wang, Zirui and Jiang, Zhiying and Zhong, Wei and Fan, Wei and Xu, Bin},
  journal={IEEE/CAA Journal of Automatica Sinica},
  volume={12},
  number={3},
  pages={502--515},
  year={2024},
  publisher={IEEE}
}

@inproceedings{liu2025dcevo,
  title={Dcevo: Discriminative cross-dimensional evolutionary learning for infrared and visible image fusion},
  author={Liu, Jinyuan and Zhang, Bowei and Mei, Qingyun and Li, Xingyuan and Zou, Yang and Jiang, Zhiying and Ma, Long and Liu, Risheng and Fan, Xin},
  booktitle={Proceedings of the IEEE/CVF Conference on Computer Vision and Pattern Recognition},
  pages={2226--2235},
  year={2025}
}

@inproceedings{cheng2025one,
  title={One model for all: Low-level task interaction is a key to task-agnostic image fusion},
  author={Cheng, Chunyang and Xu, Tianyang and Feng, Zhenhua and Wu, Xiaojun and Tang, Zhangyong and Li, Hui and Zhang, Zeyang and Atito, Sara and Awais, Muhammad and Kittler, Josef},
  booktitle={Proceedings of the IEEE/CVF Conference on Computer Vision and Pattern Recognition},
  pages={28102--28112},
  year={2025}
}

@article{tang2025mask,
  title={Mask-DiFuser: A masked diffusion model for unified unsupervised image fusion},
  author={Tang, Linfeng and Li, Chunyu and Ma, Jiayi},
  journal={IEEE Transactions on Pattern Analysis and Machine Intelligence},
  year={2025},
  publisher={IEEE}
}

@article{xu2026diff,
  title={Diff-MEF: Cross-Modal Diffusion Framework With Text Prompts and Semantic Perception for Multi-Exposure Image Fusion},
  author={Xu, Han and Huang, Yunfei and Tang, Linfeng and Ma, Jiayi and Liu, Guangcan},
  journal={IEEE Transactions on Image Processing},
  volume={35},
  pages={3186--3201},
  year={2026},
  publisher={IEEE}
}

@inproceedings{chen2025spatiotemporal,
  title={Spatiotemporal blind-spot network with calibrated flow alignment for self-supervised video denoising},
  author={Chen, Zikang and Jiang, Tao and Hu, Xiaowan and Zhang, Wang and Li, Huaqiu and Wang, Haoqian},
  booktitle={Proceedings of the AAAI Conference on Artificial Intelligence},
  volume={39},
  number={3},
  pages={2411--2419},
  year={2025}
}

@article{cai2024exploring,
  title={Exploring video denoising in thermal infrared imaging: Physics-inspired noise generator, dataset, and model},
  author={Cai, Lijing and Dong, Xiangyu and Zhou, Kailai and Cao, Xun},
  journal={IEEE Transactions on Image Processing},
  volume={33},
  pages={3839--3854},
  year={2024},
  publisher={IEEE}
}

@inproceedings{liang2021swinir,
  title={Swinir: Image restoration using swin transformer},
  author={Liang, Jingyun and Cao, Jiezhang and Sun, Guolei and Zhang, Kai and Van Gool, Luc and Timofte, Radu},
  booktitle={Proceedings of the IEEE/CVF international conference on computer vision},
  pages={1833--1844},
  year={2021}
}

@inproceedings{pan2023deep,
  title={Deep discriminative spatial and temporal network for efficient video deblurring},
  author={Pan, Jinshan and Xu, Boming and Dong, Jiangxin and Ge, Jianjun and Tang, Jinhui},
  booktitle={Proceedings of the IEEE/CVF Conference on Computer Vision and Pattern Recognition},
  pages={22191--22200},
  year={2023}
}

@article{xie2021segformer,
  title={SegFormer: Simple and efficient design for semantic segmentation with transformers},
  author={Xie, Enze and Wang, Wenhai and Yu, Zhiding and Anandkumar, Anima and Alvarez, Jose M and Luo, Ping},
  journal={Advances in neural information processing systems},
  volume={34},
  pages={12077--12090},
  year={2021}
}

@article{hu2022disguised,
  title={Disguised tailing and video surveillance with solar-powered fixed-wing unmanned aerial vehicle},
  author={Hu, Shuyan and Ni, Wei and Wang, Xin and Jamalipour, Abbas},
  journal={IEEE Transactions on Vehicular Technology},
  volume={71},
  number={5},
  pages={5507--5518},
  year={2022},
  publisher={IEEE}
}

@inproceedings{wen2024panacea,
  title={Panacea: Panoramic and controllable video generation for autonomous driving},
  author={Wen, Yuqing and Zhao, Yucheng and Liu, Yingfei and Jia, Fan and Wang, Yanhui and Luo, Chong and Zhang, Chi and Wang, Tiancai and Sun, Xiaoyan and Zhang, Xiangyu},
  booktitle={Proceedings of the IEEE/CVF Conference on Computer Vision and Pattern Recognition},
  pages={6902--6912},
  year={2024}
}

@article{liu2013intelligent,
  title={Intelligent video systems and analytics: A survey},
  author={Liu, Honghai and Chen, Shengyong and Kubota, Naoyuki},
  journal={IEEE Transactions on Industrial Informatics},
  volume={9},
  number={3},
  pages={1222--1233},
  year={2013},
  publisher={IEEE}
}

@article{xu2026iterrf,
  title={IterRF: Iterative and Interpretable Mutual Enhancement for Joint Multi-Modal Image Registration and Fusion},
  author={Xu, Han and Chen, Jiajun and Yi, Xunpeng and Li, Peili and Ma, Jiayi},
  journal={International Journal of Computer Vision},
  volume={134},
  number={10},
  pages={442},
  year={2026},
  publisher={Springer}
}

@article{chen2026breaking,
  title={Breaking Low-Light Fusion Barrier: Unsupervised Darkness and Noise-Aware Visible and Infrared Image Fusion Network},
  author={Chen, Jiajun and Xu, Han and Liu, Guangcan and Ma, Jiayi},
  journal={IEEE Transactions on Image Processing},
  volume={35},
  pages={7954--7969},
  year={2026},
  publisher={IEEE}
}

\appendix
\newpage

\section*{Technical Appendices and Supplementary Material}

\section{Network Details of Pretrained Video Diffusion Model and TD-ControlNet} 
The pretrained video diffusion backbone adopts a 3D U-Net architecture composed of alternating spatial and temporal transformer blocks (Spa.-Trans. and Tem.-Trans.). Concretely, each spatial transformer module is followed by a temporal module, forming an interleaved structure that jointly models spatial appearance and temporal dynamics. The spatial transformer modules are inherited from Stable Diffusion 2.1, consisting of residual blocks, self-attention, and cross-attention layers for 2D feature extraction. Temporal modeling is achieved by dedicated temporal convolution and temporal attention layers inserted between spatial blocks. The temporal convolution is implemented as a 1D convolution with kernel size 3 along the temporal dimension, while the temporal attention is formulated as a 1D self-attention mechanism operating across frames. To preserve the pretrained spatial generative capability, all temporal transformer modules are initialized with zero output layers and then fine-tuned on video data. Besides, a U-Net skip-connection strategy is adopted in the 3D architecture, where temporally aggregated features from the encoder are directly passed to the corresponding decoder stages. These features are fused with decoder activations via concatenation, enabling effective reuse of hierarchical spatio-temporal representations and reconstruction consistency.

For TD-ControlNet, we follow the ControlNet paradigm by duplicating the encoder and middle block of the pretrained 3D U-Net, including architectures and parameters. The conditional inputs (i.e., degraded video representations and textual degradation embeddings) are fed into this copied branch to extract conditional features. These features are then injected into the frozen 3D U-Net through additional zero-initialized convolution layers, ensuring a smooth transition from unconditional generation to conditional control. Specifically, the output of the middle block in TD-ControlNet is added to the corresponding hidden representation in the middle block of original 3D U-Net. Similarly, the encoder features from the control branch are merged with the skip-connected features produced by the encoder of the backbone network. It follows the standard ControlNet formulation, while being further extended with the proposed state-modulated degradation-aware hierarchical MoE (SDH-MoE) module. Specifically, each expert in the MoE is implemented as a convolution-based network, enabling effective modeling of local spatio-temporal patterns. The MoE adaptively refines the injected conditional signals according to both the diffusion state and degradation characteristics, thereby improving the flexibility and robustness of addressing compound degradation conditions.

\section{Degradation Parameters}
To improve the diversity of training samples and simulate realistic degradations in source videos, we randomly introduce different degradation types during data preparation.

\textbf{Gaussian Noise.}
Gaussian noise is added with a standard deviation randomly sampled from
$[40/255, 120/255]$. For each video clip, a noise level is first randomly determined and shared across the clip, while independent Gaussian noise is sampled and added to individual frames. This strategy preserves a consistent noise intensity across the video while maintaining frame-level stochastic variations, thereby better simulating realistic sensor noise.

\textbf{Stripe Noise.} It mainly occurs in infrared imaging systems. As the infrared videos in the M3SVD dataset already contain realistic stripe-noise degradation, we retain the original stripe noise provided by the dataset. Thus, no additional synthetic stripe noise is added to the M3SVD infrared videos.

\textbf{Gaussian Blur.}
Gaussian blur is generated using kernel sizes randomly selected from
$\{7, 9, 11, 13, 15, 17\}$, with the standard deviation $\sigma$ randomly sampled from the range $[0.5, 3.0]$. The sampled blur parameters are applied to the corresponding source video clip. Since the visible videos in the M3SVD dataset already contain realistic blur degradation, we directly retain the original blur characteristics of these videos without introducing additional synthetic blur.

\textbf{JPEG Compression Artifacts.}
JPEG compression artifacts are generated by JPEG encoding and decoding, with the quality factor randomly sampled from the range $[10, 50]$. To account for different temporal characteristics of compression degradation, the quality factor can either remain identical across all frames within a video clip or vary across individual frames. The former simulates temporally consistent compression artifacts, whereas the latter introduces temporally varying compression distortions that may occur in practical video acquisition and transmission scenarios.

\section{Pseudo Ground-Truth Construction}

As high-quality ground-truth fused videos are unavailable for real-world video fusion, we construct pseudo ground-truth (pseudo GT) through a multi-stage \textbf{human--metric--VLM} collaborative curation procedure. It combines coarse human inspection, quantitative metric-based ranking, and vision-language-model (VLM) verification to reduce subjective bias and improve the reliability and consistency of pseudo GT.

\textbf{i) Candidate generation.}
We first collect 6,000 aligned source-video pairs from 79 scenes, covering infrared-visible, multi-exposure, and multi-focus fusion tasks with diverse scene contents and motion patterns. For each source-video pair, the corresponding high-quality, undegraded videos are processed by three representative video fusion methods, i.e., TemCoCo, VideoFusion, and UniVF, to generate three candidate fused videos. Using multiple fusion methods helps reduce the bias introduced by any single method and provides diverse fusion priors for subsequent selection.

\textbf{ii) Human-based coarse filtering.}
We then perform coarse manual inspection to remove clearly defective samples. Specifically, candidates with severe structural distortion, ghosting artifacts, obvious temporal flickering, or motion discontinuity are discarded. Manual inspection is only used for coarse quality filtering rather than directly determining the final pseudo GT. After this stage, 5,600 source-video pairs and their corresponding candidate results are retained.

\textbf{iii) Metric-based quantitative selection.}
For each retained source-video pair, the candidate fused videos are further evaluated using both spatial and temporal fusion metrics, including MI, Qabf, tF, tSSIM, and BiSWE. Since these metrics have different numerical ranges and optimization directions, their scores are first normalized and then aggregated into an overall quality score. The candidate with the highest overall score is selected as the provisional pseudo GT. In this way, fine-grained candidate selection is determined by a unified quantitative criterion rather than subjective human preference.

\textbf{iv) VLM-assisted quality verification.}
To further identify subtle perceptual and semantic defects that may not be fully captured by handcrafted metrics, we employ the pretrained LLaVA-Video-Qwen2 model for additional verification. A fixed prompt is used to assess complementary-information preservation, structural fidelity, temporal consistency, motion continuity, and residual fusion artifacts. The VLM assigns each provisional pseudo GT a quality score from 1 to 5 together with a brief justification, and samples with scores below 2 are discarded.

After the complete human--metric--VLM curation procedure, 5,119 source--fusion video pairs are retained as the final training data. By combining human perception, quantitative evaluation, and VLM-based semantic verification, the proposed construction procedure reduces dependence on subjective judgment and any single evaluation metric, while providing more reliable and diverse fusion supervision for model training.

\section{More Qualitative Comparison Results}
\begin{figure*}[!h]
    \centering    \includegraphics[width=1\linewidth]{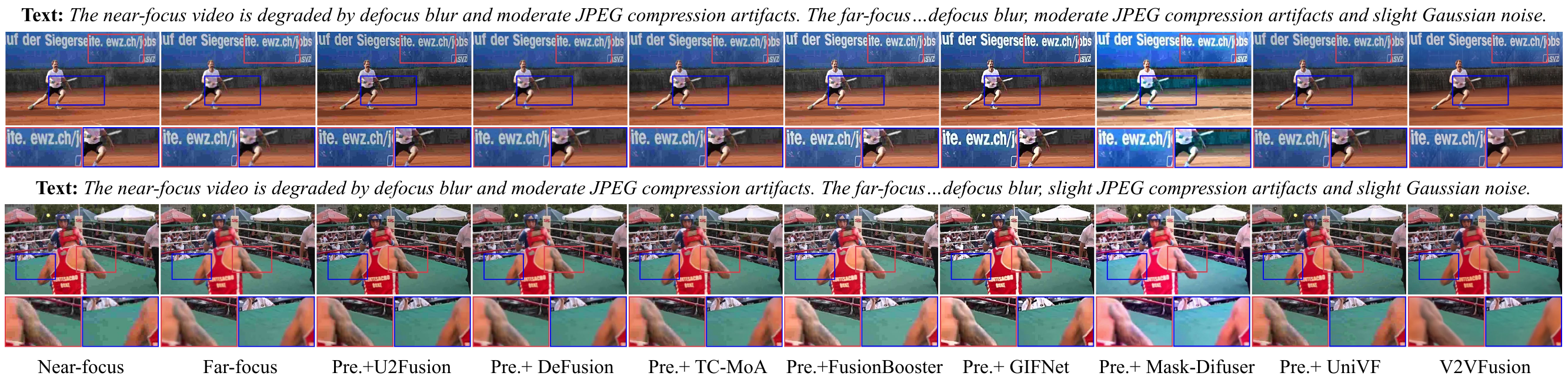}\\
    \vspace{-0.1in}
    \caption{Results of multi-focus video fusion (Pre.: pre-enhancement \cite{chen2025spatiotemporal}, \cite{cai2024exploring}, \cite{pan2023deep}, \cite{liang2021swinir}).}
    \label{app1}
\end{figure*}

\begin{figure*}[!h]
    \centering    \includegraphics[width=1\linewidth]{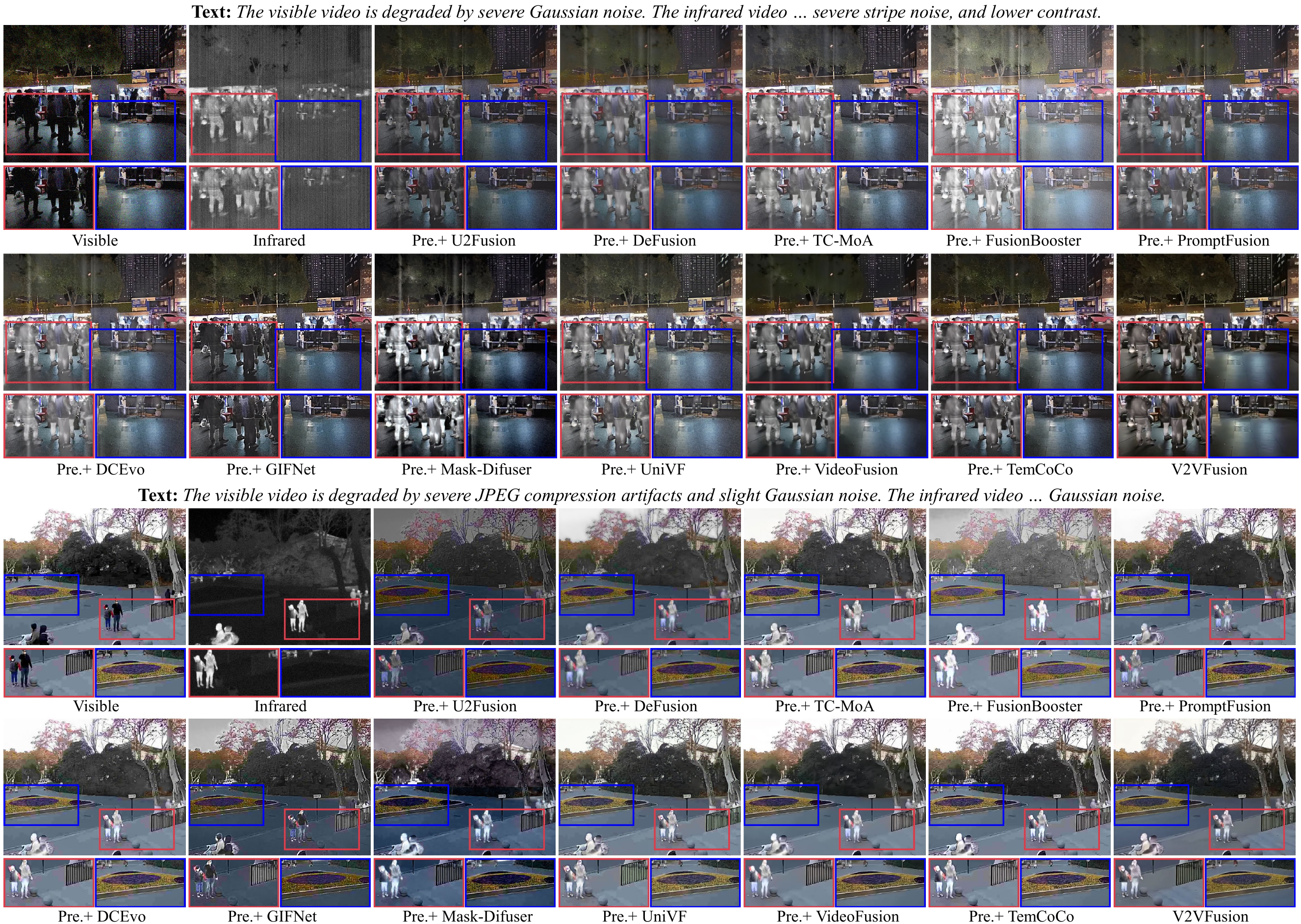}\\
    \vspace{-0.1in}
    \caption{Results of infrared-visible video fusion (Pre.: pre-enhancement \cite{chen2025spatiotemporal}, \cite{cai2024exploring}, \cite{pan2023deep}, \cite{liang2021swinir}).}
    \label{app3}
\end{figure*}

\begin{figure*}[!h]
    \centering    \includegraphics[width=1\linewidth]{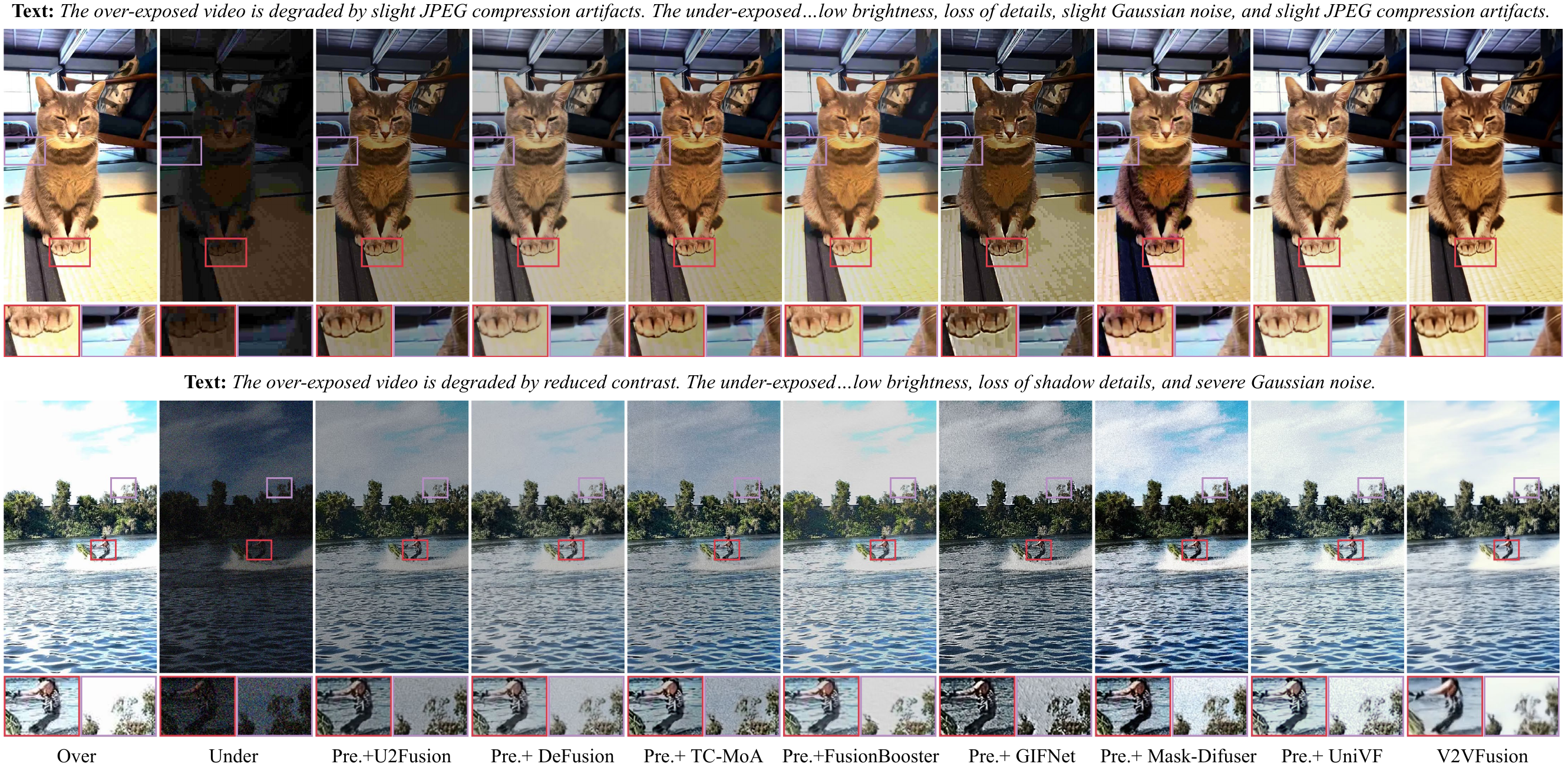}\\
    \vspace{-0.1in}
    \caption{Results of multi-exposure video fusion (Pre.: pre-enhancement \cite{chen2025spatiotemporal}, \cite{cai2024exploring}, \cite{pan2023deep}, \cite{liang2021swinir}).}
    \label{app2}
\end{figure*}

We provide additional qualitative comparison results under three video fusion scenarios. More qualitative results of multi-focus video fusion are shown in Fig.~\ref{app1}. More qualitative results of infrared-visible video fusion are shown in Fig.~\ref{app3}. More qualitative results of multi-exposure video fusion are shown in Fig.~\ref{app2}. Overall, our method achieves superior degradation suppression and produces fused videos with clearer structures, richer textures, and more stable visual quality than existing approaches.

\vspace{-0.2in}

\section{Quantitative Results on No-Reference Quality Metrics} 

We further evaluate our method using four widely adopted no-reference image quality assessment metrics, including ILNIQE, NIQE, LIQE, and CLIP-IQA. Specifically, NIQE assesses image naturalness by measuring deviations from natural scene statistics, while ILNIQE employs richer statistical features to characterize diverse visual distortions. LIQE evaluates perceptual quality by jointly modeling image content, scene semantics, and distortion characteristics through vision--language representations. CLIP-IQA leverages the semantic priors of CLIP together with quality-related textual prompts to assess overall perceptual quality.

As shown in Tab.~\ref{table_no_reference}, our method consistently achieves the best overall performance across all three fusion tasks on the four no-reference quality metrics. These results further demonstrate that our method can generate fused videos with superior perceptual quality and more natural visual appearance.
\vspace{-0.2in}
\begin{table*}[!h]
\centering
\caption{No-reference assessment on three video fusion tasks
(P.: pre-enhancement; \textcolor{red}{\textbf{red}}: optimal).}
\label{table_no_reference}
\resizebox{1\linewidth}{!}{
\renewcommand\arraystretch{1.1}
\begin{tabular}{l@{\,~}|c@{\,~~~}c@{\,~~~}c@{\,~~~}c@{\,~~~}c@{\,~~~}c@{\,~~~}c@{\,~~~}c@{\,~~~}c@{\,~~~}c@{\,~~~}c@{\,~~~}c@{\,~~}}
\toprule
\multirow{2}{*}{\textbf{Metric}}&
\multicolumn{12}{c}{\textbf{Infrared-Visible (M3SVD)}} \\
\cmidrule(lr){2-13}
& P.+U2Fus.& P.+DeFus.& P.+DCEvo& P.+FusionB.& P.+GIFNet& P.+MaskD.& P.+PromptF.& P.+TCMoA& P.+VideoFus.& P.+TemCoCo& P.+UniVF& \textbf{V2VFusion} \\
\midrule
\textbf{ILNIQE$\downarrow$}
& 30.85& 31.38& 30.52& 34.23& 32.42& 30.36& 28.87& 30.77& 28.30& 30.51& 30.15& \RH{24.02} \\

\textbf{LIQE$\uparrow$}
& 1.38& 1.46& 1.90& 1.41& 1.44& 1.74& 1.86& 1.81& 1.57
& 1.92& 1.69& \RH{2.31} \\

\textbf{CLIPIQA$\uparrow$}
& 0.37& 0.33& 0.38& 0.35& 0.38& 0.36& 0.37& 0.38& 0.38& {0.40}& 0.35& \RH{0.41} \\

\textbf{NIQE$\downarrow$}
& 5.05& 4.95& 4.46& 4.79& 5.07& {4.32}& 4.58& 4.48& 4.73& 5.45& 4.65& \RH{4.20} \\
\bottomrule
\end{tabular}
}

\resizebox{1\linewidth}{!}{
\renewcommand\arraystretch{1.2}
\begin{tabular}{l@{\,~}|c@{\,~~}c@{\,~~}c@{\,~~}c@{\,~~}c@{\,~~}c@{\,~~}c@{\,~~}c@{\,~}|c@{\,~~}c@{\,~~}c@{\,~~}c@{\,~~}c@{\,~~}c@{\,~~}c@{\,~~}c@{\,~}}
\toprule
\multirow{2}{*}{\textbf{Metric}} &
\multicolumn{8}{c|}{\textbf{Multi-Exposure (YouTube-HDR)}} &
\multicolumn{8}{c}{\textbf{Multi-Focus (DAVIS)}} \\
\cmidrule(lr){2-9} \cmidrule(lr){10-17}

& P.+U2Fus.& P.+DeFus.& P.+FusionB.& P.+GIFNet& P.+MaskD.& P.+TC-MoA& P.+UniVF& \textbf{V2VFusion}

& P.+U2Fus.& P.+DeFus.& P.+FusionB.& P.+GIFNet& P.+MaskD.& P.+TC-MoA& P.+UniVF& \textbf{V2VFusion} \\
\midrule

\textbf{ILNIQE$\downarrow$}
& 29.73& \RH{26.53}& 29.53& 32.80& 26.63& 26.90& 27.06
& \RH{26.53}

& 27.46& 29.69& 25.13& \RH{24.68}& {24.69}& 24.69& 25.37& \RH{24.68} \\

\textbf{LIQE$\uparrow$}
& 2.12& {2.53}& 2.27& 2.39& 2.33& 2.36& 2.46& \RH{2.56}

& {2.19}& 1.69& 2.18& 2.13& 1.51& 2.08& 2.11& \RH{2.21} \\

\textbf{CLIPIQA$\uparrow$}
& {0.50}& {0.50}& 0.45& {0.50}& 0.45& 0.49
& {0.50}& \RH{0.51}

& {0.47}& 0.44& 0.46& 0.43& 0.37& 0.45
& 0.45& \RH{0.50} \\

\textbf{NIQE$\downarrow$}
& 4.68& 3.72& 4.25& 4.82& {3.65}& {3.65}& 3.74
& \RH{3.64}

& 4.68& 4.67& 4.93& 4.61& {4.30}& 4.31& 4.37
& \RH{4.29} \\
\bottomrule
\end{tabular}
}
\end{table*}

\section{Qualitative Results of Ablation Study}
\begin{figure*}[!h]
    \centering    \includegraphics[width=1.0\linewidth]{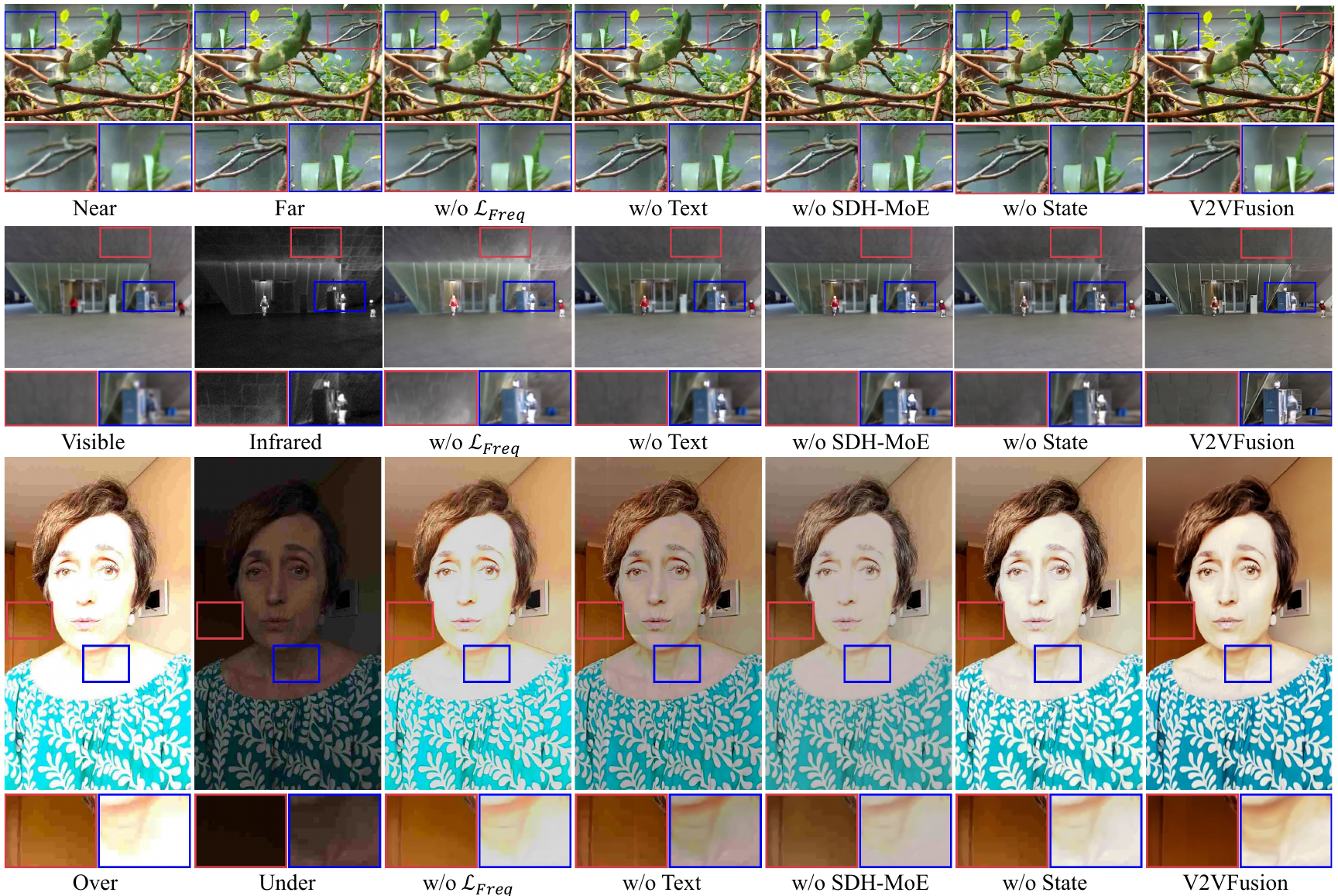}
    \vspace{-0.2in}
    \caption{Qualitative result comparisons of the ablation study.}
    \label{ablation}
\end{figure*}

We provide qualitative comparisons of the ablation study in Fig.~\ref{ablation}. 
Specifically, we analyze several variants by removing key components, including the frequency reconstruction loss (w/o $\mathcal{L}_{Freq}$), textual degradation descriptions (w/o $\text{Text}$), the proposed SDH-MoE module (w/o $\text{SDH-MoE}$), and the state-modulated mechanism in SDH-MoE (w/o $\text{State}$). 
The results show that removing any of these components leads to noticeable degradation in both degradation suppression and fusion quality, further verifying the effectiveness of each component in our method.

\section{Fast-Motion Fusion Evaluation}
Video fusion requires not only preserving informative spatial details from different modalities, but also maintaining temporally consistent motion of dynamic objects across consecutive frames. While static regions should retain fine structural and texture details, moving targets should exhibit coherent appearance and motion trajectories without introducing ghosting, misalignment, or temporal artifacts. These requirements are more challenging in fast-motion scenarios, where large inter-frame displacement and rapid temporal variation significantly increase the difficulty of temporal modeling.

\begin{figure*}[!h]
    \centering
    \includegraphics[width=1.0\linewidth]{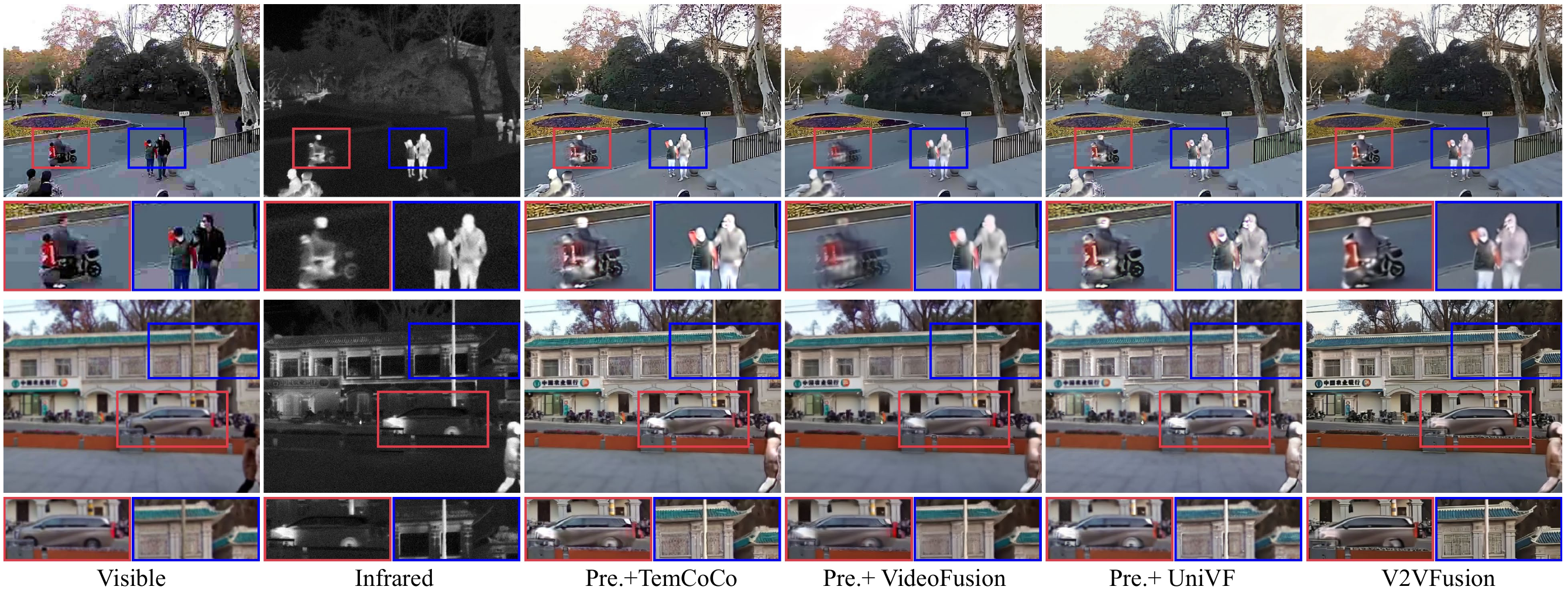}
    \vspace{-0.2in}
    \caption{Qualitative comparison of visible-infrared video fusion under fast-motion scenarios.}
    \label{fast_motion}
\end{figure*}

To evaluate the robustness under such challenging conditions, we conduct qualitative comparisons on sequences containing fast-moving objects. As shown in Fig.~\ref{fast_motion}, existing video fusion methods tend to suffer from noticeable structural distortions and motion-related artifacts in rapidly moving regions. In comparison, our method can preserve salient targets in the infrared modality while retaining detailed appearance information in the visible modality. Meanwhile, it maintains more stable target structures across consecutive frames, demonstrating stronger robustness to fast motion in video fusion.

\section{Long-Video Fusion Evaluation}

To further evaluate the temporal robustness of different video fusion methods on long video sequences, we conduct experiments with video lengths of 90, 120, and 150 frames. We report four temporal consistency metrics, including tF, tLP, tSSIM, and BiSWE. As shown in Tab.~\ref{tab_long_video}, the performance of competing methods generally degrades as the video length increases, indicating that maintaining temporal consistency becomes increasingly challenging for long-term video fusion. In contrast, our method exhibits substantially better robustness across different sequence lengths.
\vspace{-0.1in}
\begin{table*}[!h]
\centering
\caption{Comparisons on long videos with different numbers of frames.}
\label{tab_long_video}
\resizebox{\linewidth}{!}{
\renewcommand{\arraystretch}{1.1}
\begin{tabular}{l|cccc|cccc|cccc}
\toprule
\multirow{2}{*}{\textbf{Method}}
& \multicolumn{4}{c|}{\textbf{90 Frames}}
& \multicolumn{4}{c|}{\textbf{120 Frames}}
& \multicolumn{4}{c}{\textbf{150 Frames}} \\
\cmidrule(lr){2-5}
\cmidrule(lr){6-9}
\cmidrule(lr){10-13}

& \textbf{tF}$\downarrow$
& \textbf{tLP}$\downarrow$
& \textbf{tSSIM}$\uparrow$
& \textbf{BiSWE}$\downarrow$
& \textbf{tF}$\downarrow$
& \textbf{tLP}$\downarrow$
& \textbf{tSSIM}$\uparrow$
& \textbf{BiSWE}$\downarrow$
& \textbf{tF}$\downarrow$
& \textbf{tLP}$\downarrow$
& \textbf{tSSIM}$\uparrow$
& \textbf{BiSWE}$\downarrow$ \\
\midrule

TemCoCo
& 0.93 & 0.28 & 0.97 & 2.23
& 0.95 & 0.31 & 0.95 & 2.35
& 1.32 & 0.36 & 0.93 & 3.02 \\

VideoFusion
& 0.64 & 0.23 & 0.97 & 1.13
& 0.73 & 0.29 & 0.94 & 1.35
& 1.02 & 0.32 & 0.92 & 2.79 \\

UniVF
& 1.40 & 0.27 & 0.96 & 2.52
& 1.56 & 0.30 & 0.95 & 2.61
& 1.72 & 0.35 & 0.92 & 2.67 \\

\textbf{V2VFusion}
& \textbf{\RH{0.49}} & \textbf{\RH{0.14}} & \textbf{\RH{0.99}} & \textbf{\RH{0.85}}
& \textbf{\RH{0.51}} & \textbf{\RH{0.16}} & \textbf{\RH{0.99}} & \textbf{\RH{0.87}}
& \textbf{\RH{0.53}} & \textbf{\RH{0.18}} & \textbf{\RH{0.98}} & \textbf{\RH{0.90}} \\
\bottomrule
\end{tabular}
}
\end{table*}


\section{Limitations} 
Although the proposed framework demonstrates strong performance on multiple degradation-conditional fusion scenarios, the range of degradations considered in this work is primarily focused on common types (e.g., blur, Gaussian noise, stripe noise, and JPEG artifacts). Degradations that may arise from specific sensor characteristics or environmental conditions during imaging are not yet included. This limitation mainly stems from the scarcity of large-scale real-world multi-source video fusion datasets containing diverse degradation types. In future work, we plan to address this limitation by constructing a more comprehensive multi-source video fusion dataset with diverse real-world degradations. Such datasets would facilitate more reliable evaluation and promote the development of more robust video fusion methods under complex real-world conditions.

\newpage
\section*{NeurIPS Paper Checklist}

\begin{enumerate}

\item {\bf Claims}
    \item[] Question: Do the main claims made in the abstract and introduction accurately reflect the paper's contributions and scope?
    \item[] Answer: \answerYes{}
    \item[] Justification: The abstract and introduction of the paper clearly state the claims made, including the  contributions made in the paper and important assumptions and limitations.

\item {\bf Limitations}
    \item[] Question: Does the paper discuss the limitations of the work performed by the authors?
    \item[] Answer: \answerYes{}
    \item[] Justification: The paper discusses the limitations of the work performed by the authors, providing a balanced view of the study’s scope and potential areas for improvement.

\item {\bf Theory assumptions and proofs}
    \item[] Question: For each theoretical result, does the paper provide the full set of assumptions and a complete (and correct) proof?
    \item[] Answer: \answerYes{}
	\item[] Justification: All the theorems, formulas, and proofs in the paper have been numbered and cross-referenced, and all assumptions have been clearly stated or referenced in the statement of any theorems.

    \item {\bf Experimental result reproducibility}
    \item[] Question: Does the paper fully disclose all the information needed to reproduce the main experimental results of the paper to the extent that it affects the main claims and/or conclusions of the paper (regardless of whether the code and data are provided or not)?
    \item[] Answer: \answerYes{} 
	\item[] Justification: The paper has fully disclosed all the information needed to reproduce the main experimental results of the paper to the extent that it affects the main claims and conclusions of the paper.

\item {\bf Open access to data and code}
    \item[] Question: Does the paper provide open access to the data and code, with sufficient instructions to faithfully reproduce the main experimental results, as described in supplemental material?
    \item[] Answer: \answerYes{}
    \item[] Justification: We will open access to the code after the paper is accepted.

\item {\bf Experimental setting/details}
    \item[] Question: Does the paper specify all the training and test details (e.g., data splits, hyperparameters, how they were chosen, type of optimizer, etc.) necessary to understand the results?
    \item[] Answer: \answerYes{}
	\item[] Justification: We have specified all the training and test details necessary to understand the results.

\item {\bf Experiment statistical significance}
    \item[] Question: Does the paper report error bars suitably and correctly defined or other appropriate information about the statistical significance of the experiments?
    \item[] Answer: \answerYes{}
	\item[] Justification: The paper reports error bars and other appropriate information about the statistical significance of the experiments, ensuring the reliability and validity of the results.

\item {\bf Experiments compute resources}
    \item[] Question: For each experiment, does the paper provide sufficient information on the computer resources (type of compute workers, memory, time of execution) needed to reproduce the experiments?
    \item[] Answer: \answerYes{}
	\item[] Justification: The paper provides sufficient information on the computer resource for experiments.
    
\item {\bf Code of ethics}
    \item[] Question: Does the research conducted in the paper conform, in every respect, with the NeurIPS Code of Ethics \url{https://neurips.cc/public/EthicsGuidelines}?
    \item[] Answer: \answerYes{}
	\item[] Justification: The research conducted in the paper conforms in every respect with the NeurIPS Code of Ethics, ensuring ethical standards are upheld throughout the study.

\item {\bf Broader impacts}
    \item[] Question: Does the paper discuss both potential positive societal impacts and negative societal impacts of the work performed?
    \item[] Answer: \answerYes{}
	\item[] Justification: The paper discusses both potential positive societal impacts and negative societal impacts of the work performed, providing a comprehensive evaluation of the study’s broader implications.
    
\item {\bf Safeguards}
    \item[] Question: Does the paper describe safeguards that have been put in place for responsible release of data or models that have a high risk for misuse (e.g., pretrained language models, image generators, or scraped datasets)?
    \item[] Answer: \answerYes{}
	\item[] Justification: The paper describes safeguards that have been put in place for the responsible release of data or models that have a high risk for misuse, ensuring that appropriate measures are taken to mitigate potential risks.

\item {\bf Licenses for existing assets}
    \item[] Question: Are the creators or original owners of assets (e.g., code, data, models), used in the paper, properly credited and are the license and terms of use explicitly mentioned and properly respected?
    \item[] Answer: \answerYes{}
	\item[] Justification: The creators or original owners of assets used in the paper are properly credited, and the license and terms of use are explicitly mentioned and properly respected.

\item {\bf New assets}
    \item[] Question: Are new assets introduced in the paper well documented and is the documentation provided alongside the assets?
    \item[] Answer:\answerYes{}
	\item[] Justification: The new assets introduced in the paper are well documented, and the documentation is provided alongside the assets.

\item {\bf Crowdsourcing and research with human subjects}
    \item[] Question: For crowdsourcing experiments and research with human subjects, does the paper include the full text of instructions given to participants and screenshots, if applicable, as well as details about compensation (if any)? 
    \item[] Answer: \answerNA{}
	\item[] Justification: The paper does not involve crowdsourcing nor research with human subjects.

\item {\bf Institutional review board (IRB) approvals or equivalent for research with human subjects}
    \item[] Question: Does the paper describe potential risks incurred by study participants, whether such risks were disclosed to the subjects, and whether Institutional Review Board (IRB) approvals (or an equivalent approval/review based on the requirements of your country or institution) were obtained?
    \item[] Answer: \answerNA{}
	\item[] Justification: The paper does not involve crowdsourcing nor research with human subjects.

\item {\bf Declaration of LLM usage}
    \item[] Question: Does the paper describe the usage of LLMs if it is an important, original, or non-standard component of the core methods in this research? Note that if the LLM is used only for writing, editing, or formatting purposes and does not impact the core methodology, scientific rigorousness, or originality of the research, declaration is not required.
    \item[] Answer: \answerNA{}
	\item[] Justification: The core method development in the paper does not involve LLMs as any important, original, or non-standard components.

\end{enumerate}
\end{document}